\documentclass[lettersize,journal]{IEEEtran}
\usepackage{amsmath,amsfonts}
\usepackage{algorithm}
\usepackage{array}
\usepackage[caption=false,font=normalsize,labelfont=sf,textfont=sf]{subfig}
\usepackage{textcomp}
\usepackage{stfloats}
\usepackage{url}
\usepackage{amsmath}
\usepackage{verbatim}
\usepackage{graphicx}
\usepackage[nocompress]{cite}
\usepackage{multirow}
\usepackage{rotating}
\usepackage{makecell}
\usepackage{ragged2e}
\usepackage{color}
\usepackage{xcolor}
\definecolor{citecolor}{HTML}{3498DB}
\definecolor{linkcolor}{HTML}{E74C3C}
\usepackage[pagebackref=false,breaklinks=true,colorlinks,bookmarks=false,citecolor=blue,linkcolor=linkcolor]{hyperref}
\usepackage{amssymb}
\usepackage{bbding}
\usepackage{booktabs}
\usepackage{colortbl}
\usepackage{threeparttable}
\usepackage{physics}
\usepackage{bm}
\usepackage[english]{babel} 
\usepackage[normalem]{ulem}
\addto\captionsenglish{}
\usepackage{ltxtable}
\begin{document}

\title{Embodied Co-Design for Rapidly Evolving Agents: Taxonomy, Frontiers, and Challenges}

\author{Yuxing Wang, Zhiyu Chen, Tiantian Zhang, Qiyue Yin, Yongzhe Chang, Zhiheng Li, Liang Wang, Xueqian Wang\\ 
\thanks{Yuxing Wang, Zhiyu Chen, Tiantian Zhang, Yongzhe Chang, Zhiheng Li, and Xueqian Wang are with the Shenzhen International Graduate School (SIGS), Tsinghua University, Shenzhen 518055, China. E-mail: \{wyx20, zhiyu-ch21\}@mails.tsinghua.edu.cn; zhhli@tsinghua.edu.cn; \{zhang.tt, changyongzhe, wang.xq\}@sz.tsinghua.edu.cn.}
\thanks{Qiyue Yin and Liang Wang are with the New Laboratory of Pattern Recognition (NLPR), Institute of Automation, Chinese Academy of Sciences, Beijing 100190, China. E-mail: \{qyyin, wangliang\}@nlpr.ia.ac.cn.}
}



\maketitle

\begin{abstract}
Brain-body co-evolution enables animals to develop complex behaviors in their environments. Inspired by this biological synergy, \emph{embodied co-design} (ECD) has emerged as a transformative paradigm for creating intelligent agents—from virtual creatures to physical robots—through the joint optimization of morphology and control. This integrated approach facilitates richer environmental interactions and enhances task performance. In this survey, we provide a systematic overview of recent advances in ECD. We first formalize the concept of ECD and position it within related fields. We then introduce a hierarchical taxonomy in which a lower layer breaks down agent design into three fundamental components—\emph{controlling brain}, \emph{body morphology}, and \emph{task environment}—while an upper layer integrates these components into four major ECD frameworks: \emph{bi-level}, \emph{single-level}, \emph{generative}, and \emph{open-ended}. This taxonomy provides a unified lens through which we distill insights from over one hundred recent studies. We further review notable benchmarks, datasets, and applications in both simulated and real-world scenarios. Finally, we discuss significant challenges and offer insights into promising future research directions. A project associated with this survey has been established at \url{https://github.com/Yuxing-Wang-THU/SurveyBrainBody}.
\end{abstract}  


\section{Introduction} 
\IEEEPARstart{E}{mbodied} intelligence (EI) seeks to integrate cognitive capabilities into autonomous machines, enabling them to learn from and adapt to complex environments~\cite{howard2019evolving}. Situated at the intersection of artificial intelligence and robotics, EI has achieved significant advancements in fusing perception, decision-making, and actuation, empowering various robots to tackle previously intractable challenges, such as dynamic locomotion~\cite{lee2020learning,zhang2024whole}, end-to-end navigation~\cite{zheng2024towards,schumann2024velma}, and dexterous manipulation~\cite{chi2023diffusion,huang2023voxposer,wang2024dexcap}, to name just a few. 

In traditional EI, an agent’s morphology (\emph{body}) is typically hand-designed and treated as a fixed component of the \emph{environment}, while only its controller (\emph{brain}) is optimized for task execution. However, this separation presents clear limitations. A pre-specified morphology can constrain sensing and actuation, thereby narrowing the space of achievable behaviors. In other words, when morphological design prioritizes mechanical optimality without considering how it will be controlled, the burden of achieving coordinated behavior shifts almost entirely to the controller. This imbalance often creates a bottleneck in task performance and raises a fundamental question: \emph{How should we design the optimal form of an embodied agent to effectively perform tasks in its environment?}

\begin{figure}[t]
\centering
\includegraphics[width=0.485\textwidth]{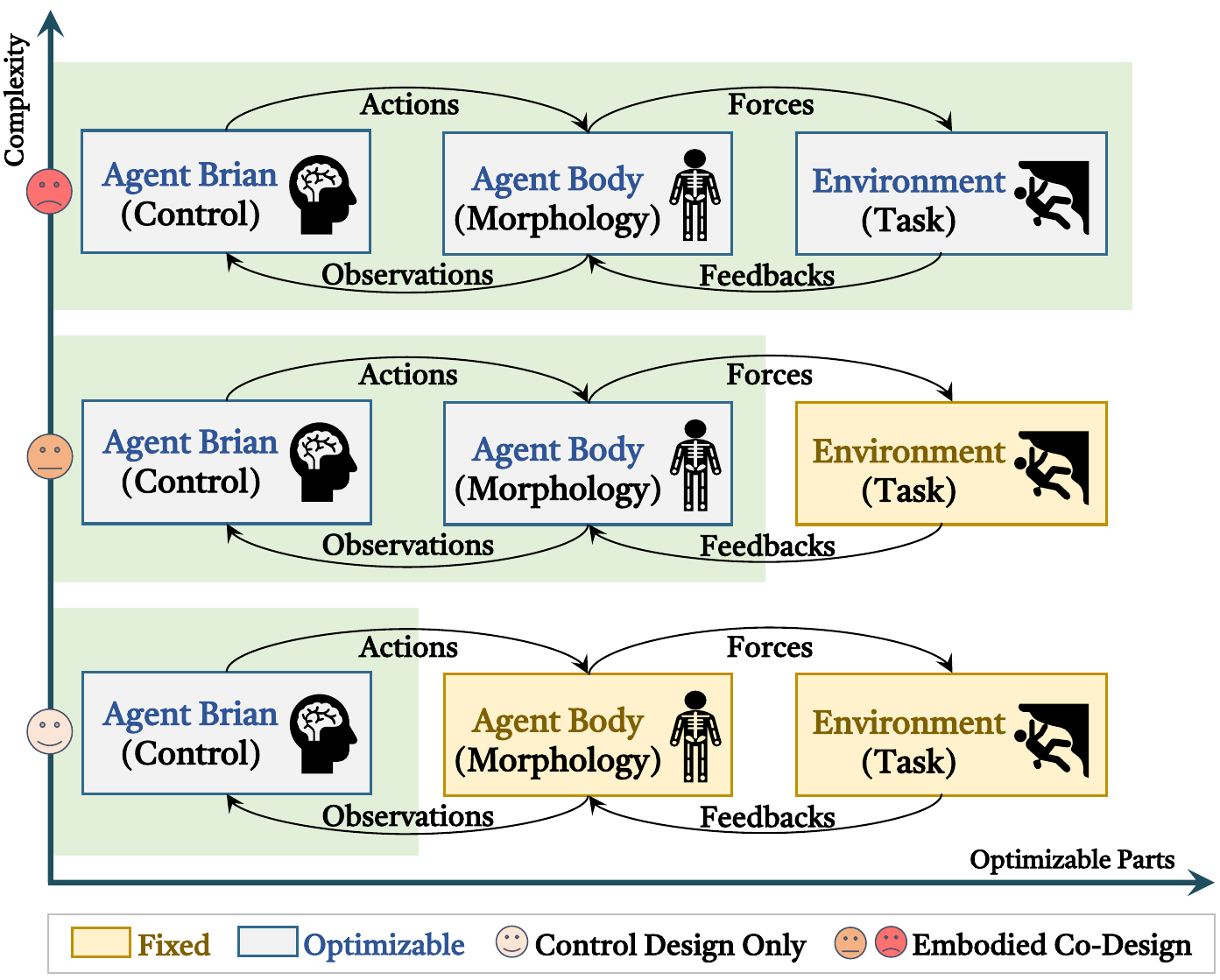}
\caption{Differences between control design only and embodied co-design.}
\label{fig_1}
\end{figure}
\IEEEpubidadjcol
The concept of morphological computation~\cite{paul2006morphological,muller2017morphological,pfeifer2009morphological,pfeifer2006body,mertan2024no} posits that a well-designed body can simplify computations typically assigned to the brain, thereby facilitating more efficient behaviors. This perspective introduces a \emph{brain-body dynamic}: changes in morphology necessitate corresponding adaptations in control to exploit newly acquired capabilities, while environmental feedback can, in turn, guide further refinement of morphology. These synergies have motivated various efforts to improve the current agent design procedure. As a result, \emph{embodied co-design} (ECD), an integrated framework that co-optimizes an agent's morphology with its control mechanism to enhance task performance, becomes a critical topic in EI.

ECD was initially explored in the \emph{Evolutionary Robotics} (ER) community~\cite{harvey2005evolutionary,silva2016open}, where both robot morphology and behavior are jointly optimized using \emph{Evolutionary Algorithms} (EAs)~\cite{harvey2005evolutionary}. However, this topic remains emergent for EI, particularly as agents and their tasks continue to increase in complexity and diversity. Compared to controller-only optimization, ECD introduces a vast, often combinatorial design space for morphology-control configurations, accompanied by fragile interdependencies. Specifically, evaluating a candidate morphology typically requires re-training the controller from scratch, as even minor morphological changes can invalidate previously learned controllers. Variations in environmental conditions, design constraints, and task objectives produce heterogeneous feedback signals, further complicating the co-design process (Fig.~\ref{fig_1}). Consequently, ECD problems are highly non-convex and computationally expensive, underscoring the growing need for automated methodologies instead of traditional manual engineering. 

\emph{Motivation:} Recent breakthroughs in domains such as evolutionary algorithms~\cite{davis2023subtract,bergonti2024co,ferigo2025totipotent}, reinforcement learning (RL)~\cite{yuan2022transformact,wang2023curriculum, dongleveraging2024,lu2025bodygen}, differentiable simulation~\cite{xu2021accelerated,ma2021diffaqua,Strgar-RSS-24}, and large models~\cite{gao2025vlmgineer,fang2025robomore,song2025laser} have led to a surge of novel ECD approaches that rapidly reshape the capabilities of co-design systems. However, their assumptions, representations, computational pipelines, evaluation benchmarks, and limitations have not been systematically reviewed, making it difficult for researchers to develop a coherent understanding of the ECD landscape. Our goal is to clarify the conceptual foundations, consolidate frontier developments, highlight emerging trends, and identify open challenges that will shape the next generation of research in EI.

\emph{Taxonomy:} To enhance readability for researchers across disciplines, we propose a two-layer hierarchical taxonomy in Section~\ref{tax}. The lower layer structures ECD around three core dimensions: \emph{controlling brain} (Section~\ref{brain}), \emph{body morphology} (Section~\ref{body}), and \emph{task environment} (Section~\ref{environment}). Building upon this foundation, the upper layer categorizes existing ECD methods into four major frameworks: \emph{bi-level} (Section~\ref{bi-level}), \emph{single-level} (Section~\ref{single-level}), \emph{generative} (Section~\ref{genertive}), and \emph{open-ended} (Section~\ref{open}), with further subcategories identified within each. This taxonomy establishes a common vocabulary and provides a conceptual map for ECD, facilitating cross-domain understanding. It also encourages researchers to explore beyond their primary areas. For instance, specialists in RL for control optimization may discover how their methodologies align with a specific ECD framework, in conjunction with previously unconsidered morphological evolution, thereby opening pathways for new collaborations.  

\emph{Scope:} Embodied co-design bridges robotics, artificial intelligence, control theory, and cognitive science. Guided by our proposed taxonomy, we analyze the design spaces and representation methods for morphology, control, and task environments, with a particular emphasis on how these components are coordinated within various co-design frameworks. Rather than exhaustively reviewing any specific algorithmic subfield (e.g., RL for control optimization) or engineering detail (e.g., 3D printing for morphological fabrication), this work provides an organizational perspective that clarifies the conceptual and technical interrelations among different ECD works.

\emph{Related Works:} Most prior surveys either provide broad overviews of embodied intelligence~\cite{long2025survey,duan2022survey,sun2024comprehensive,liu2025embodied} or focus on isolated subfields such as embodied learning~\cite{liang2025large}, manipulation~\cite{li2025survey}, and navigation~\cite{xiong2025sensing}. These works emphasize sensorimotor integration, while attention to ECD has been largely limited. To our knowledge, partial reviews of ECD exist in specific domains, including soft robotics~\cite{zhao2024exploring,chen2020design,stroppa2024optimizing}, humanoid robots~\cite{liu2025embracing,yue2025toward}, evolutionary robotics~\cite{doncieux2014beyond,prabhu2018survey, alattas2019evolutionary,pandey2022accessible,strgar2024evolution}, and artificial life~\cite{lai2021virtual,casadei2023artificial,ha2022collective}. However, these reviews lack a structured taxonomy for systematically comparing ECD methods; they do not incorporate recent advances, benchmarks, and emerging applications.

\emph{Features:} Unlike previous works, our survey is the \emph{\textbf{first}} to provide a systematic, method-centric analysis of ECD across various agent types, optimization paradigms, and application domains. Rather than treating morphology as a secondary concern or focusing on isolated subfields, we, for the \emph{\textbf{first}} time, introduce a two-layer hierarchical taxonomy that explicitly unifies the design process of embodied agents, thus enabling principled comparisons across various ECD methods. We further incorporate recent advances—including differentiable simulation, quality-diversity search, latent and foundation-model-based generation—together with benchmarks, datasets, and real-world testbeds that have been largely absent from related reviews. By combining conceptual unification with practical evaluation criteria and open challenges, this survey offers both a coherent theoretical framework and an up-to-date technical reference for the ECD community.

\emph{Contributions:} The main contributions of our survey are summarized below:
\begin{enumerate}

\item We conceptualize ECD from three complementary perspectives: bi-level optimization, sequential decision-making, and model-based optimization (Section~\ref{back}), providing a multifaceted view of the ECD problem.

\item We contribute a hierarchical taxonomy (Section~\ref{tax}) that organizes co-design spaces and identifies key design choices, offering a unified framework for understanding the fundamental components of ECD. 

\item We present a systematic technical review (Section~\ref{methods}) of major advances in ECD over the past five years, analyzing representative methods and their core principles.

\item We survey simulation platforms, datasets, and real-world testbeds in ECD research (Section~\ref{benapp}), highlighting their capabilities and representative demonstrations.

\item We identify current bottlenecks in ECD and outline promising research directions (Section~\ref{futu}) that may benefit a broad range of communities and help advance the state of the art.
\end{enumerate}

\section{Background: Formalizing Embodied Co-Design}\label{back}
\subsection{Problem Definition}\label{pd}
\subsubsection{Embodied Agents} An embodied agent, such as a robot, can be characterized by its \emph{body morphology} and \emph{controlling brain}. The \emph{body morphology} $\omega \in \Omega$ defines the agent’s physical interface with its environment, encompassing structural topology, material properties, and the spatial configuration of sensors and actuators. The \emph{controlling brain} $\pi\in\Pi$ maps observed states to action commands, driving the body to exert forces that interact with the \emph{task environment}. These forces induce state transitions governed by the physical dynamics $F\in\mathcal{F}$, which, in turn, generate new sensory inputs. Through this sensorimotor loop, the agent learns to produce inherently goal-directed behaviors, with its performance evaluated using a fitness function $R\in\mathcal{R}$. 

\subsubsection{Bi-Level Optimization View} ECD naturally has a bi-level structure, in which one optimization task is nested within another~\cite{sinha2017review,liu2021investigating}. The goal is to identify both the optimal morphology $\omega ^{*}$ and the optimal control policy $\pi^{*}$, leading to the following formulation:
\begin{equation}
\label{eq1}
\begin{split}
&\omega ^{*}=\underset{\omega\in\Omega}{argmax}\,R(\pi^{*},\omega)\\
    &s.t.\quad \pi^{*}=\underset{\pi\in\Pi}{argmax}\,R(\pi,\omega)\\
\end{split}
\end{equation} 
Here, the upper level searches for the optimal morphology, while the lower level trains a controller for the specified morphology to maximize the agent's performance. Notably, this bi-level characteristic is also observable in other deep learning problems, such as \emph{Neural Architecture Search} (NAS)~\cite{liu2021survey}, where the objective is to jointly optimize both the network structure (analogous to \emph{morphology} in ECD) and its parameters (analogous to \emph{control} in ECD). 

\subsubsection{Sequential Decision-Making View} While the bi-level formulation captures the hierarchical structure inherent in ECD, it fails to specify the causal connection between morphology and control. To bridge this gap, we formulate ECD as a single-level \emph{Markov Decision Process} (MDP)~\cite{wang2022deep}. Following the sequential workflow commonly adopted in engineering practice, we divide the overall co-design process into two main stages: the design stage $\mathcal{D}$ and the control stage $\mathcal{C}$. As a result, the ECD problem can be represented as a $9$-tuple $\left \langle \mathcal{S_D},\mathcal{S_C},\mathcal{A_D},\mathcal{A_C}, \mathcal{F_D},\mathcal{F_C},\mathcal{R_D},\mathcal{R_C},\gamma \right \rangle$, where $\gamma \in(0,1]$ denotes the reward discount factor, with the remaining elements corresponding to the relevant design spaces. For clarity, explicit representations of constraints—such as morphological or task-specific constraints—are omitted and incorporated implicitly within their respective design spaces. 

In this formulation, the overall state space comprises two components: the design state $s_{t}^{d}\in\mathcal{S_D}$, which governs morphology creation, and the control state $s_{t}^{c}\in\mathcal{S_C}$, which governs interactions with the environment. The corresponding action space includes design actions $a_{t}^{d}\in\mathcal{A_D}$ and control actions $a_{t}^{c}\in\mathcal{A_C}$. Under an episodic framework, a morphology is initially generated as $\omega = s_{T_d}^{d}$ by a design policy $\pi_{d}(a_{t}^{d}|s_{t}^{d})$, which executes a sequence of design actions guided by the design transition dynamics $F_{d}(s^{d}_{t+1}|s^{d}_{t}, a^{d}_{t})$ from the initial design state $\omega_0=s^{d}_{0}$ to the final design step $T_d$. In the subsequent control stage, a policy $\pi_{c}(a_{t}^{c}|s_{t}^{c},\omega)$ conditioned on $\omega$ directs the generated morphology to perform tasks, following the environmental transition dynamics $F_c(s^{c}_{t+1}|s^{c}_t, a^{c}_t,\omega)$. Based on the aforementioned notations, the single-level objective $R$ of ECD is now expressed as:
\begin{equation}
\label{eq22}
\begin{split}
R=\mathbb{E}_{\pi_{d},\pi_{c}} \left[\sum_{t=0}^{T_d}\gamma ^{t}r^{d}_{t}+\sum_{t=T_d+1}^{\infty}\gamma ^{t}r^{c}_{t} \right]
\end{split}
\end{equation} 
where $r^{d}_{t}\in\mathcal{R_D}$ and $r^{c}_{t}\in\mathcal{R_C}$ denote the rewards obtained during the design and control stages, respectively. By leveraging effective gradient approximation techniques in RL~\cite{schulman2017proximal}, both the design policy and the control policy can be iteratively updated through trial-and-error interactions.

\subsubsection{Model-Based Optimization View} It is noteworthy that, in the perspectives discussed above, the environmental transition model $F_c(s^{c}_{t+1}|s^{c}_t, a^{c}_t,\omega)$ is typically treated as a black box. However, as discussed in Section~\ref{phy}, many ECD approaches construct $F_c$ either as a differentiable simulator~\cite{hu2019chainqueen,du2021diffpd,hu2019difftaichi} or as a learned dynamics model~\cite{navez2025modeling}. Conceptually, the transition can be written as an abstract function $s^{c}_{t+1} = F_c(s^{c}_t, a^{c}_t,\omega) $, where $a^{c}_t$ denotes a vector of control signals, such as the joint torques applied during that time-step. Thus, given an initially parameterized morphology $\omega$ and a sequence of actions potentially derived from a control policy, the simulator unrolls a trajectory to a terminal control state $s^{c}_{T_c}$ and computes a task loss $\mathcal{L}$ that depends on this final state (e.g., distance traveled or final velocity~\cite{ma2021diffaqua,xu2021accelerated}). Since the simulator or learned model is differentiable with respect to environmental dynamics, morphological parameters, and control variables, gradients can propagate through the entire trajectory via the chain rule:
\begin{align}
     \pdv{\mathcal{L}}{s^{c}_t} = \left(\pdv{\mathcal{L}}{s^{c}_{t+1}}\right)\left(\pdv{F_c}{s^{c}_t}\right) \\
    \pdv{\mathcal{L}}{a^{c}_t} = \left(\pdv{\mathcal{L}}{s^{c}_{t+1}}\right)\left(\pdv{F_c}{a^{c}_t}\right) \\
    \pdv{\mathcal{L}}{\omega} = \left(\pdv{\mathcal{L}}{s^{c}_{t+1}}\right)\left(\pdv{F_c}{\omega}\right)
\end{align}
\label{eq:backward}
This model-based formulation facilitates end-to-end, gradient-based optimization for inverse design in ECD, providing substantial advantages in sample efficiency.  
\subsection{Philosophical Foundations}\label{phi}
\emph{Embodied Cognition} has emerged from a sustained dialog between philosophical reflection and empirical science. Early philosophical discussions often treated cognition as an abstract, disembodied process largely independent of physical embodiment~\cite{descartes2016meditations}. This perspective was increasingly challenged in the mid-twentieth century, notably by Turing~\cite{turing2021computing}, who argued that intelligence should arise through learning and adaptation, envisioning machines that grow and learn like children. Subsequent theoretical and empirical advances have further strengthened the critique of brain-centric perspectives. Clark~\cite{clark1998being} emphasized the centrality of sensorimotor processes, arguing that bodily interactions with the environment both constrain and enable cognitive capacities. Building on this foundation, Pfeifer and Scheier proposed that the body is not merely an instrument for perception and action, but rather an essential component of cognitive processing~\cite{pfeifer2006body,pfeifer2001understanding}. 

\emph{Morphological Computation} recognizes that an agent’s physical body can inherently perform computations that simplify control and perception, rather than serving solely as a passive structure governed by a controller~\cite{paul2006morphological,muller2017morphological,pfeifer2009morphological,wilson2013embodied,koehl1996does}. A classic illustration is the passive dynamic walker~\cite{collins2005efficient}, which achieves stable locomotion solely through its morphology, without any software control. However, such examples also reveal critical limitations: passive solutions typically support only a narrow set of behaviors and are effective only under constrained environmental conditions. This introduces a fundamental \emph{brain-body trade-off}~\cite{xie2025morphology,hoffmann2017simple,rosendo2017trade,zhang2025co}: simple morphologies allow lightweight controllers for basic tasks, whereas more expressive morphologies often require sophisticated control mechanisms to fully exploit their capabilities in complex environments. A central goal in ECD, therefore, is to achieve an effective balance between morphological computation (as quantified in Section~\ref{metric}) and controller-driven (informational) computation.

\emph{Evolutionary Theory} suggests that brains and bodies co-evolve through adaptive processes driven by environmental pressures. Central to this theory is \emph{Natural Selection}, originally articulated by Darwin~\cite{bonner1988evolution}, which posits that traits that enhance an organism’s survival and reproduction are more likely to be passed on to future generations. Over time, this process shapes increasingly specialized and efficient morphologies and behaviors, improving the organism’s fit to its environment. Complementing this view, the \emph{Baldwin Effect} links evolution and learning by suggesting that an organism’s ability to learn can accelerate evolutionary adaptation~\cite{jelisavcic2019lamarckian,luo2023enhancing,simpson1953baldwin}. Another corollary of evolution is that more complex environments stimulate the development of more capable brain–body systems~\cite{mahon2008critical,smith2005development,smith2005cognition}. Recent empirical findings from~\cite{gupta2021embodied} have confirmed these theories, showing that agents evolved in progressively more challenging terrains develop morphologies that enhance downstream learning efficiency.

Our survey will position various ECD methods within these foundational principles, illustrating how ECD integrates these insights to design agents that are both structurally sound and behaviorally adept.

\subsection{Related Topics}\label{rela}
\emph{Evolutionary Robotics} (ER) originally focused on the evolution of robot behaviors~\cite{bongard2013evolutionary,scheper2016behavior}. Over time, the scope expanded to include the evolutionary design of both robotic hardware and control policies. A seminal contribution is Sims’ work~\cite{sims1994evolving}, which evolves 3D virtual creatures by jointly optimizing their graph-based morphologies and neural controllers. Contemporary ECD builds upon these insights while extending beyond traditional ER by incorporating a broader array of representation and learning techniques to enhance scalability, generalization, and applicability.

\emph{Artificial Life} (A-Life) investigates the co-evolution of morphology and control, aiming to recreate life-like processes in virtual environments~\cite{moreno2005agency,aguilar2014past}. While A-Life shares the central concept of brain–body–environment coupling with ECD, it prioritizes the exploration of biological phenomena such as self-organization~\cite{gershenson2020self}, adaptation~\cite{bedau2003artificial}, and evolutionary dynamics~\cite{ito2016population} in nature, rather than engineering practical embodied agents with specific functionalities.

\emph{Computational Robot Design} leverages algorithmic tools to automate the synthesis of robots, typically following a sequential design process. This approach formulates complex constrained optimization problems and employs techniques such as topology optimization~\cite{sigmund2000topology,sigmund2013topology} and CAD integration~\cite{tian2022assemble, willis2022joinable} to address issues related to structural feasibility, material selection, electronics integration, and fabrication constraints. Control, however, is often simplified or regarded as a secondary component. In contrast, ECD is a broader concept that tightly integrates morphology and control within a unified optimization loop, aiming to generate not only feasible body designs but also adaptive, task-capable behaviors.

\emph{Robot Learning} aims to equip robots with the ability to autonomously acquire perceptual, decision-making, and action skills, rather than relying solely on hand-engineered solutions~\cite{kroemer2021review}. It is closely connected to ECD, as learning-based control optimization serves as a central mechanism for solving tasks such as locomotion, manipulation, and navigation. Recent advances in robot learning have integrated model-based approaches~\cite{grandia2023perceptive}, reinforcement learning~\cite{matsuo2022deep}, large-model-based reasoning~\cite{xiao2025robot}, imitation learning~\cite{hussein2017imitation}, and self-supervised methods~\cite{jeong2020self} to enhance high-dimensional sensorimotor loops. However, most robot learning frameworks assume a fixed morphology. ECD generalizes this setting by treating morphology and control as interdependent variables and optimizing them jointly.

\section{Taxonomy: Decomposing Embodied Co-Design}\label{tax}
\begin{figure*}[t]
\centering
\includegraphics[width=0.99\textwidth]{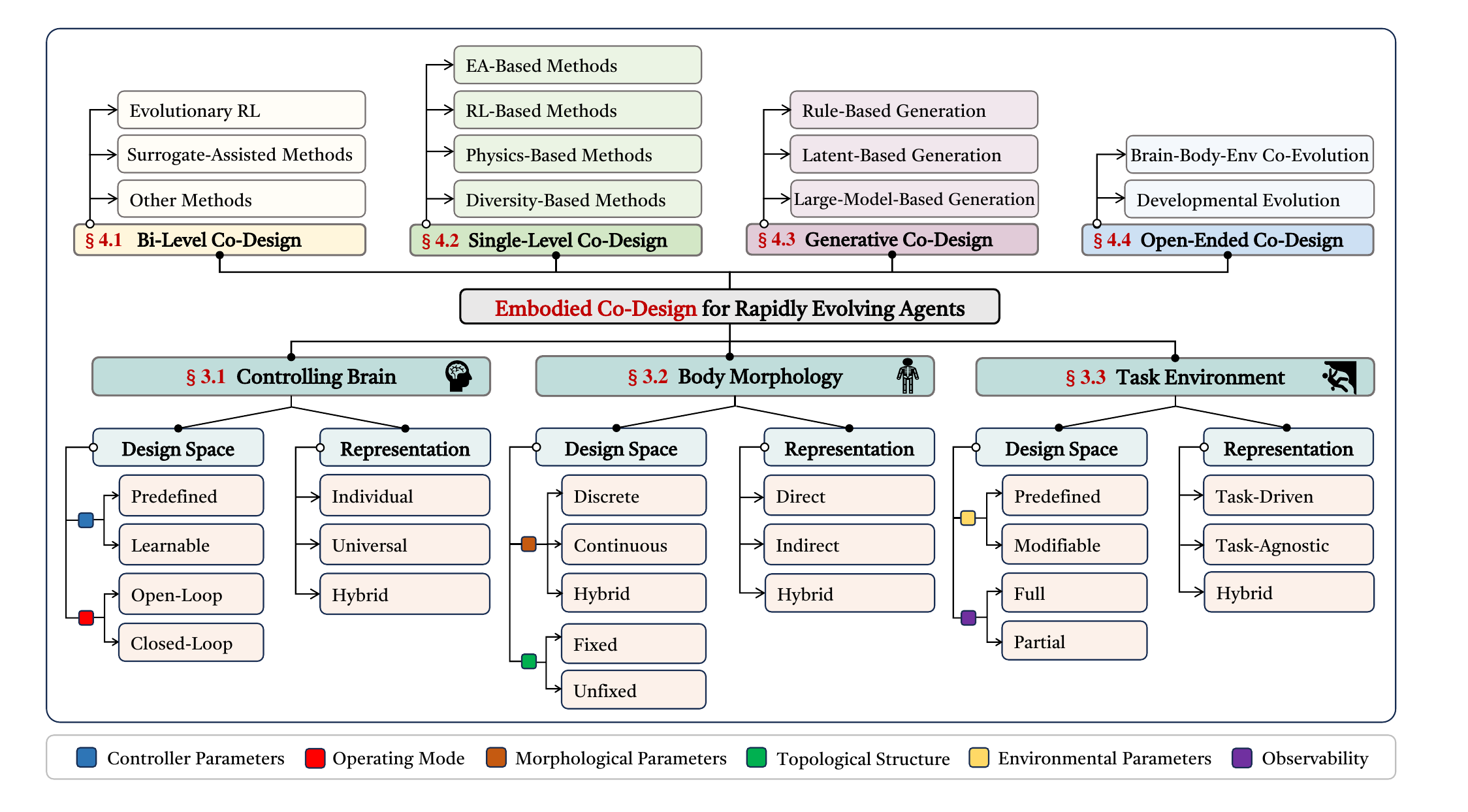}
 \captionsetup{justification=raggedright, singlelinecheck=false}
\caption{An overview of the embodied co-design landscape including foundational elements and co-design frameworks.}
\label{fig_2}
\end{figure*}

ECD operates within an expansive workspace shaped by the interplay of control, morphology, and task environment. As shown in Fig.~\ref{fig_2}, our taxonomy organizes these fundamental elements into a lower layer, while the upper layer defines four ECD frameworks, each with sub-categories that guide the co-design process. In the following sections, we detail each component, highlight recent advances, discuss common constraints, summarize key evaluation metrics, then illustrate a representative ECD pipeline using a prototype.

\subsection{Controlling Brain}\label{brain}
The controlling brain enables an agent to process sensory information and generate appropriate motor responses for body control, thereby closing the sensorimotor loop.

\subsubsection{Design Space} The design space of controlling brain encompasses both the ways by which controllers are parameterized and the modes in which they operate.

\emph{Controller Parameters} can be \emph{\textbf{predefined}}—such as action primitives~\cite{iii2021taskagnostic}, fixed waveform generators~\cite{cheney2013unshackling}, and hand-crafted gaits~\cite{kobayashi2024computational}—or they can be fully \emph{\textbf{learnable}}, optimized through methods such as trajectory optimization, reinforcement learning, and imitation learning. Modern ECD approaches typically employ learnable controllers that enable adaptation to varying morphologies and support flexible, real-time improvements in behavior.

\emph{Operating Mode} defines the fundamental strategy by which a control system executes its commands and manages its actions. The two primary modes of control are \emph{open-loop} and \emph{closed-loop}. \emph{\textbf{Open-loop}} controllers execute predefined action sequences without incorporating real-time sensory feedback; they are efficient and effective for simple, predictable tasks but cannot compensate for disturbances or failures during execution. In contrast, \emph{\textbf{closed-loop}} controllers continuously incorporate sensory feedback to adapt their actions, offering greater robustness and generality at the cost of increased computational demands and more expressive policy representations. In embodied co-design, the closed-loop operating mode is typically favored.

\subsubsection{Representation} The controller representation defines the mapping between the control policy and the agent's physical structure. Classical approaches adopt either \emph{individual} or \emph{universal} representations, while recent research explores \emph{hybrid} variants that combine their strengths.

\emph{\textbf{Individual}} representations directly bind the controller architecture or dimensionality to a specific morphology. Examples include low-level PID controllers with per-joint gains that account for the inertia and stiffness of specific body segments~\cite{ang2005pid,borase2021review}, as well as \emph{Model Predictive Control} (MPC) when an accurate dynamic model is available~\cite{schwenzer2021review,carron2019data}. Another example is \emph{Central Pattern Generators} (CPGs), which mirror a robot’s limb topology to produce coordinated rhythmic motion~\cite{bliss2012central,deshpande2023deepcpg}. Higher-level neural controllers, such as \emph{Multi-Layer Perceptrons} (MLPs), adopt input/output sizes tied to a robot’s sensors and actuators. Individualized controllers are suitable for \emph{topology-fixed} agents that share identical morphological structures but differ solely in their kinematic or dynamic parameters. This specialization, however, reduces transferability; even minor morphological changes—such as the addition of a joint or modifications to limb geometry—typically necessitate restructuring the controller. Consequently, per-agent representations struggle to generalize across heterogeneous state and action spaces.

\emph{\textbf{Universal}} representations aim to generalize across diverse morphologies. Representing a body as a kinematic graph naturally invites the use of \emph{Graph Neural Networks} (GNNs)~\cite{munikoti2023challenges}; their message-passing mechanisms enable a single policy to generalize across graphs with varying node counts, thereby supporting scalable control across different robotic embodiments. Representative GNN-based controllers include Nervernet~\cite{wang2018nervenet}, DGN~\cite{pathak2019learning}, Whitman et al.~\cite{whitman2023learning}, SMP~\cite{huang2020one}, MAT~\cite{li2024mat}, MG~\cite{pan2025morphology}, and GCNT~\cite{ijcai2025p972}. Subsequent research in cognitive science has further demonstrated that modularity can enhance control generalization~\cite{schilling2024modularity}. To capture the long-range dependencies that GNNs inherently struggle to address, transformer-based controllers incorporate global attention over all body parts. This concept has been explored in numerous embodied AI studies, such as Amorpheus~\cite{kurin2020my}, MetaMorph~\cite{gupta2022metamorph}, SWAT~\cite{hong2021structure}, AnyMorph~\cite{trabucco2022anymorph}, Xiong et al.~\cite{xiong2023universal}, Body Transformer~\cite{sferrazza2024body}, HeteroMorpheus~\cite{hao2024heteromorpheus}, and URMA~\cite{bohlinger2024one}. Beyond ECD, universal controllers have also shown strong utility in transfer learning~\cite{chen2018hardware,peng2018sim,cho2024meta}, meta-learning~\cite{nagabandi2018learning,ghadirzadeh2021bayesian}, and cross-embodiment learning~\cite{yang2024pushing, shafiee2024manyquadrupeds,wang2024sparse,chen2024mirage,patel2025get,mertan2024towards,feng2023genloco,luo2024moral}.

\emph{\textbf{Hybrid}} representation seeks to balance specialization and generalization by decomposing the controller or control mechanism into a shared, morphology-agnostic module that encodes universal motor and perceptual priors, and a morphology-specific module that adapts outputs to specific embodiments~\cite{bjelonic2023learning,chen2023evolving,luck2020data,luck2021robot,sliacka2023co,sun2023co}. Hybrid representations can be seamlessly integrated into various pipelines, such as pretraining-finetuning~\cite{yue2025toward,jin2025evolutionary} and meta-learning~\cite{belmonte2022meta}, to facilitate cross-morphology generalization and accelerate adaptation during the ECD process.

\subsubsection{Discussion}\label{tax:control}
Training a dedicated controller can provide an accurate estimate of a morphology’s potential by fully exploiting its passive dynamics and affordances. Empirical results from~\cite{kvalsund2022centralized,mertan2023modular,chiappa2022dmap,luo2023comparison} have demonstrated that well-designed controllers can enhance morphological diversity and mitigate premature convergence. However, they require more comprehensive evaluation efforts to ``diagnose'' whether performance gains arise from enhanced controller capabilities or advantageous morphological adaptations. This raises an important question: \emph{must one optimize a controller to evaluate morphology?} Recent work in ECD suggests that the answer may be no. For instance, TAME~\cite{iii2021taskagnostic} introduces an information-theoretic, controller-free fitness function that measures a morphology’s intrinsic ability to explore its environment and produce predictable state transitions using only random action primitives.

\subsection{Body Morphology}\label{body}
Body morphology refers to the physical structure of an embodied agent. It is a set of properties that define the agent’s mechanical form, material composition, and physical interaction capabilities.
\subsubsection{Design Space}
The morphological design space specifies the allowable parametric and structural variations. 

\emph{Morphological Parameters} are commonly categorized as \emph{discrete}, \emph{continuous}, and \emph{hybrid}. \emph{\textbf{Discrete}} parameters typically arise in modular systems where agents are composed of a finite set of building blocks, topologies, or design configurations. \emph{\textbf{Continuous}} parameters govern real-valued attributes such as link lengths, shapes, and material properties. More general embodied systems adopt a \emph{\textbf{hybrid}} scheme in which certain choices, such as whether to attach a limb, are discrete, whereas attributes like limb length are continuous. This formulation balances structural flexibility with optimization tractability, enabling scalable searches over high-dimensional morphologies.

\emph{Topological Structure} further defines the scope of morphological exploration. In \emph{\textbf{fixed}} topology settings, such as those involving bipedal~\cite{ha2019reinforcement}, quadrupedal~\cite{belmonte2022meta}, or manipulator skeletons~\cite{cai2023task2morph}, optimization is constrained to parametric variations within a predefined template. This approach ensures stability and facilitates hardware deployment; however, it restricts the discovery of novel morphologies. In contrast, \emph{\textbf{unfixed}} topology settings permit arbitrary structural modifications, ranging from voxel-based~\cite{bhatia2021evolution} to graph-based agents~\cite{yuan2022transformact} capable of growing, pruning, or reconfiguring their skeletons. While enabling open-ended exploration, these approaches also raise concerns regarding manufacturability.

\subsubsection{Representation}
Morphological representation refers to the encoding of an agent’s body. Three paradigms are commonly employed: \emph{direct}, \emph{indirect}, and \emph{hybrid}.

\emph{\textbf{Direct}} encodings explicitly specify the agent's morphological attributes—such as link lengths, joint locations, voxel occupancy, and actuator strengths—using vectors~\cite{luck2020data}, matrices~\cite{bhatia2021evolution, sun2023co}, densities~\cite{sato2023topology,kobayashi2024computational,sato2025computational}, or trees~\cite{gupta2021embodied,jelisavcic2019lamarckian,harada2024lamarckian}. Although highly interpretable and precise, direct schemes often result in high-dimensional search spaces that hinder scalability for complex morphologies.

\emph{\textbf{Indirect}} encodings compress the agent's morphological information into compact descriptors or generative rules, which are then expanded into complete morphologies through developmental or decision-making processes. Examples include L-systems~\cite{veenstra2022effects}, generative grammars~\cite{zhao2020robogrammar}, \emph{Neural Cellular Automata} (NCA)~\cite{veenstra2020different}, \emph{Compositional Pattern-Producing Networks} (CPPNs)~\cite{cheney2013unshackling}, and reinforcement learning-based decision-making~\cite{yuan2022transformact,wang2023curriculum}. Indirect encodings typically yield smoother fitness landscapes~\cite{thomson2024understanding}, facilitate the escape from morphological local optima, and produce modular, symmetric morphologies with enhanced transferability.

\emph{\textbf{Hybrid}} encodings aim to combine the strengths of both approaches by explicitly parameterizing part of the morphology while generating the remainder procedurally or via learned models~\cite{dongleveraging2024,ringel2025text2robot,saito2024effective,le2024improving,mertan2025controller}. This design facilitates both global exploration and local refinement. However, it also introduces the challenge of balancing the search across heterogeneous spaces while maintaining coherence between explicit and generated components.

\subsubsection{Discussion}
The trade-off between direct and indirect morphological encodings remains an open question in ECD. Prior work~\cite{pigozzi2023morphology} has reported that direct encodings often yield faster learning and more consistent convergence. In contrast, indirect encodings generate more diverse, spatially coherent, and globally varying morphologies; however, their impact on learning performance is less predictable. Comparative studies~\cite{komosinski2001comparison,veenstra2017evolution} further demonstrate that indirect encodings provide advantages during early exploration due to their compactness and genetic reuse, while direct encodings may excel during later optimization stages. These insights underscore the potential of hybrid strategies that begin with indirect encodings for broad exploration and subsequently transition to direct encodings for fine-grained refinement.

\subsection{Task Environment}\label{environment}
The task environment defines the external conditions under which an agent operates. 

\subsubsection{Design Space} The design space of task environment specifies whether environmental parameters remain fixed or vary during co-design and how feedback signals are delivered.

\emph{Environmental Parameters} are typically categorized into \emph{predefined} and \emph{modifiable} types. In \emph{\textbf{predefined}} settings, all environmental attributes, such as terrain roughness and object placement, are fixed, resulting in stable and repeatable dynamics. However, this specialization often results in brittle solutions that degrade under unmodeled perturbations or when deployed beyond training conditions. In contrast, treating the environment as \emph{\textbf{modifiable}}—for example, by sampling from a distribution of scenario parameters, introducing curriculum-based difficulty, or employing non-stationary task sequences—encourages agents to generalize more effectively.

\emph{Observability} is a fundamental property of the task environment, as it determines the fidelity and completeness of the information an agent can perceive about the underlying system state. In a \emph{\textbf{fully}} observable setting, the agent's observations constitute a sufficient statistic of the true environment state. This condition simplifies learning and planning, as observed in classical control and various reinforcement learning tasks. Conversely, \emph{\textbf{partial}} observability occurs when agents receive incomplete~\cite{schaff2022soft}, noisy~\cite{bjelonic2023learning}, or local per-body/per-joint information~\cite{wang2023curriculum,schaff2022n,yuan2022transformact}, which decouples immediate observations from the global state. In such cases, the agent must maintain an internal belief state or memory to integrate temporal observations. The degree of observability, therefore, inherently influences an agent's perceptual and control mechanisms; it is also a critical axis for benchmarking an agent's robustness and generalization in real-world scenarios.

\subsubsection{Representation}
The environmental representation defines the objectives, constraints, and evaluative (reward) signals that guide the optimization dynamics of ECD. We broadly categorize these representations into three types: \emph{task-driven}, \emph{task-agnostic}, and \emph{hybrid}. 

\emph{\textbf{Task-driven}} representations specify explicit performance goals—such as locomotion speed, energy efficiency, manipulation accuracy, and task completion—providing direct reward signals that align with predefined objectives, whether single or multiple. These representations support efficient optimization but may bias the search toward narrow solutions. In contrast, \emph{\textbf{task-agnostic}} representations employ objective functions that do not encode a specific downstream task. Instead, they promote properties such as behavioral diversity, exploration, novelty, and morphological richness (see Section~\ref{metric}). By decoupling environment representation from explicit task rewards, these approaches encourage broader coverage of the design space and can reveal unanticipated yet beneficial behaviors. \emph{\textbf{Hybrid}} approaches integrate both paradigms, combining explicit task specifications with auxiliary task-agnostic signals. This joint formulation balances directed optimization with exploratory pressure and may serve as a more effective driving force for embodied co-design.

\subsubsection{Discussion}\label{multi}
The task environment shapes, and is shaped by, the agent’s morphology and control. Prior studies indicate that robots co-designed on uneven terrain tend to develop more stable, balanced, and limb-rich morphologies, whereas flat terrains favor simpler forms~\cite{miras2020environmental,miras2019effects,spanellis2021environment, corucci2018evolving}. Moreover, morphologies designed in complex, multi-task environments exhibit stronger transfer to novel tasks than those developed in simpler settings~\cite{gupta2021embodied,wang2023preco}. Note that multi-task scenarios introduce additional \emph{brain–body trade-offs}, as morphological adaptations that improve one task may degrade performance on others. Consequently, multi-task ECD seeks to cultivate broadly capable brain–body pairs by leveraging the shared structures across tasks.

\subsection{Co-Design Constraints} 
Despite their significance, many ECD approaches still operate in largely unconstrained settings, which can lead to overfitting. Specifically, morphological search may yield designs that are unnecessarily complex for a single task or, conversely, result in trivial solutions with poor transferability. Therefore, principled regularization is essential for guiding the search toward balanced brain-body pairs. 

\emph{Morphological constraints}, such as structural stability, material availability, joint range-of-motion, and actuation limits, can be explicitly incorporated into the design process. Common strategies include enforcing symmetry priors~\cite{dong2023symmetry,gupta2021embodied} and penalizing manufacturing time, cost, and complexity in the objective function, as in monotone co-design~\cite{censi2015mathematical,zardini2021co,wilhelm2023constraint}. \emph{Behavioral constraints}, such as safety, actuator saturation, and collision avoidance, can further limit the feasible behavior space and hinder the overuse of simulators.

\subsection{Co-Design Frameworks}
Based on the fundamental elements mentioned above, we categorize ECD approaches into four frameworks—(1) bi-level co-design, (2) single-level co-design, (3) generative co-design, and (4) open-ended co-design—because these categories capture the most significant differences in \emph{\textbf{how}} algorithms coordinate morphology, control, and, optionally, the task environment during optimization (Section~\ref{methods}). 

The bi-level/single-level split distinguishes between hierarchical (nested) and unified (joint) optimization loops, thereby separating methods based on evaluation cost, convergence behavior, and morphological diversity. The generative category isolates approaches that rely on learned or rule-based priors to generate candidate bodies, emphasizing the trade-offs between sample efficiency and design freedom. Open-ended co-design, finally, groups methods that promote continual innovation through the co-evolution of task environments and agents. Together, these axes establish a design space for ECD methods that can guide the development of new embodied agents and their customization for specific applications.

\subsection{Evaluation Metrics}\label{metric}
In this section, we summarize representative metrics for evaluating ECD approaches, each of which quantifies different aspects of co-designed agent performance.
\subsubsection{Task Performance} 
Task performance remains the primary application-level metric, as it directly measures goal achievement. Evaluating this aspect is costly during morphology search because it typically requires complete policy optimization and extensive rollout-based testing.

\subsubsection{Morphological Computation} 
Morphological computation (MC) quantifies the contribution of body–environment dynamics to state transitions, independent of control mechanisms. A common formalization uses conditional mutual information:
\begin{equation}
\label{eq5}
    \mathrm{MC} \;=\; I\big(s^{c}_{t+1}; s^{c}_t \mid a^{c}_t\big)
\end{equation}
which measures how informative the previous control state $s^c_t$ is about the next state $s^c_{t+1}$ when the action $a^c_t$ is known. A large MC indicates that morphology and environment strongly drive state transitions. Although recent studies have directly quantified and optimized MC in physical robots~\cite{chandiramani2024improving,ghazi2016evaluating}, most practical applications rely on simplified or surrogate estimators~\cite{zahedi2013quantifying,muller2017morphological}.
\subsubsection{Empowerment} 
Similar to MC, empowerment measures the maximum influence an agent can exert on future states through its actions~\cite{klyubin2005empowerment,leibfried2019unified}, and it is widely used in reinforcement learning exploration~\cite{houthooft2016vime}. Formally, the empowerment value for a control state $s^{c}_{t}$ is defined as the \emph{channel capacity} between the action $a^{c}_{t}$ and the next state $s^{c}_{t+1}$:
\begin{equation}
    \mathcal{E}(s) = \max_{\pi} I(s^{c}_{t+1};a^{c}_{t}|s^{c}_{t})
\end{equation}
where $\pi$ is the empowerment maximizing policy. In ECD, high empowerment indicates that a morphology can effectively translate diverse control signals into predictable outcomes, making downstream control easier~\cite{iii2021taskagnostic}.

\subsubsection{Quality-Diversity Score}
As discussed in Section \ref{diver}, diversity-based ECD methods aim to discover a wide repertoire of high-performing morphologies and behaviors. Accordingly, the Quality-Diversity (QD) score measures both coverage and performance over a discretized \emph{Morphological Descriptor} (MD) space, where each cell corresponds to a distinct morphological niche. For an archive with $N$ cells, the QD score is computed as the sum of the task returns $\mathcal{R}_{i}$ of the best individual in each occupied cell:
\begin{equation}
    QDScore =  {\textstyle \sum_{i=1}^{N}}\mathcal{R}_i
\end{equation}
\subsubsection{Complexity}
Complexity captures the cost associated with both the controller and the morphology. Controller complexity can be measured using architecture-agnostic metrics, such as FLOPs, which estimate computational load, latency, and energy consumption~\cite{xie2025morphology}. Morphological complexity can be quantified via component counts (e.g., links, joints, degrees of freedom) or structural descriptors (e.g., connectivity, material heterogeneity, and symmetry).

\subsubsection{Robustness} Robustness evaluates an agent’s resilience to perturbations, including morphological damage (e.g., locked joints, sensor faults) and environmental variations~\cite{powers2021good}. Robust agents maintain high performance and enable rapid adaptation, reflecting their inherent fault tolerance.
 
In summary, these metrics provide a multidimensional evaluation framework for ECD approaches, helping researchers enhance their understanding of how physical structures and controllers computationally contribute to behavior.

\subsection{Case Analysis}
At the bottom of Fig. \ref{fig_3}, we use \emph{BipedalWalker} agent \footnote{\url{https://gymnasium.farama.org/environments/box2d/bipedal_walker/}} in the OpenAI Gymnasium platform as a working example to summarize the common pipeline of ECD. 
\begin{itemize}
    \item \emph{Controlling Brain:} The controller is a \emph{learnable} multilayer perceptron (MLP) that receives joint angles, velocities, and ground-contact signals as inputs and outputs joint torques (\emph{individual} representation), forming a standard \emph{closed-loop} controller architecture.
    \item \emph{Body Morphology:} The agent has two legs, each composed of two segments, yielding a \emph{continuous}, \emph{topology-fixed} design space. Morphology is \emph{directly} encoded by an eight-dimensional vector specifying the width and height of each leg segment. The sizes are constrained to values within 75\% of the leg sizes in the original bipedal walker environment~\cite{wang2019poet,stensby2021co}.
    \item \emph{Task Environment:} The agent operates within a \emph{predefined}, \emph{fully observable} terrain. The goal is to maximize the forward distance traveled, which is \emph{task-driven} and characterized as a single-objective task reward function without constraints.
    \item \emph{Co-Design Framework:} A \emph{bi-level co-design} framework is employed. Morphologies are randomly initialized, and the outer-loop morphological optimization is performed by a genetic algorithm, while the inner-loop controller optimization is carried out using reinforcement learning.
\end{itemize}

\section{Methods: Synthesizing Embodied Co-Design}\label{methods}
\begin{figure*}[t]
\centering
\includegraphics[width=0.99\textwidth]{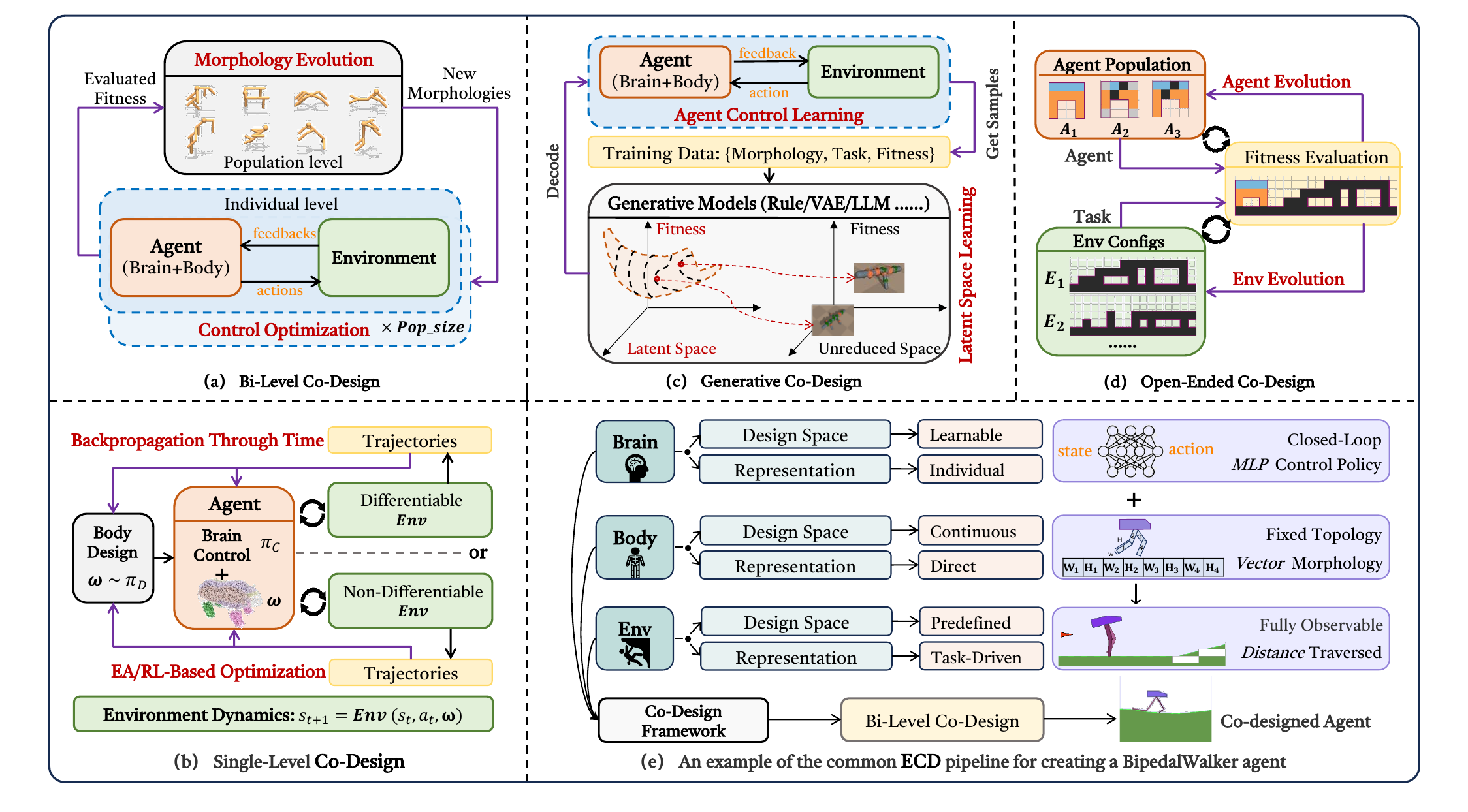}
\caption{An overview of four embodied co-design frameworks. Example images of co-designed agents from (a) to (e) are adapted from DERL~\cite{gupta2021embodied}, Softzoo~\cite{wang2023softzoo}, EvolutionGym~\cite{bhatia2021evolution}, RoboGrammar~\cite{zhao2020robogrammar}, and~\cite{ha2019reinforcement}.}
\label{fig_3}
\end{figure*}
In this section, we elaborate on four ECD frameworks and organize them into sub-categories following the taxonomy outlined in Section~\ref{tax}. For each sub-category, we analyze its core methodological components and highlight recent advances. An overview of these frameworks is presented in Fig.~\ref{fig_3}.

\subsection{Bi-Level Co-Design}\label{bi-level}
We start by introducing the bi-level co-design framework, which is the most widely adopted ECD approach. In bi-level co-design, nested optimization is employed: the inner loop optimizes the control policy for each morphology, while the outer loop explores the morphological space based on the performance evaluations of the morphologies, as illustrated in Fig.~\ref{fig_3} (a). We organize recent work along this line into three subtopics: 1) \emph{\textbf{Evolutionary Reinforcement Learning}}, 2) \emph{\textbf{Surrogate-Assisted Methods}}, and 3) \emph{\textbf{Other Methods}}.  

\subsubsection{Evolutionary Reinforcement Learning}
In this method, the inner loop for control policy optimization typically employs reinforcement learning (RL)~\cite{singh2022reinforcement,wang2022deep}, while the outer loop for morphology evolution utilizes various evolutionary algorithms (EAs).

Two major classes of RL methods are used for policy optimization: \emph{model-based} and \emph{model-free} approaches. Model-free RL learns a policy directly from interactions without explicitly modeling transition dynamics. Representative algorithms include proximal policy optimization (PPO)~\cite{schulman2017proximal}, trust region policy optimization (TRPO)~\cite{schulman2015trust}, deep deterministic policy gradient (DDPG)~\cite{lillicrap2015continuous}, twin delayed DDPG~\cite{fujimoto2018addressing}, and soft actor-critic (SAC)~\cite{haarnoja2018soft}. In contrast, Model-based RL leverages either a provided or learned dynamics model to refine the control policy. Representative frameworks include Mb-Mf~\cite{nagabandi2018neural}, PETS~\cite{chua2018deep}, GPS~\cite{levine2013guided}, and PILCO~\cite{deisenroth2011pilco}. Model-based approaches are more sample-efficient but may struggle in complex domains where accurate dynamics modeling is harder than direct policy learning.

EAs are effective for exploring high-dimensional morphology spaces. In the outer loop, \emph{evolutionary operators} determine how new morphologies are generated, thereby shaping both the breadth and depth of exploration~\cite{carvalho2024role}. In \emph{discrete} design spaces, the \emph{mutation} operator typically adds or removes structural elements—e.g., joints, limbs, modules, and kinematic branches—or modifies material properties. In \emph{continuous} design spaces—including link lengths, joint stiffness, and mass distribution—Gaussian perturbations are commonly used. The \emph{crossover} operator further expands the search space by recombining structural components or genotype encodings from multiple parents. Notably, generative models, such as \emph{Large Language Models} (LLMs), have been used as high-level mutation operators to propose structured, human-interpretable edits based on prompts~\cite{song2025laser}, further expanding the operator toolbox for morphology search.

By alternating between individual-level controller learning and population-level morphology evolution, ERL methods reflect Darwinian evolution, enhanced by a Baldwinian learning mechanism~\cite{luo2022effects,li2023evaluation}. A representative example of this line is \emph{Neural Graph Evolution} (NGE)~\cite{wang2019neural}, which evolves graph-structured morphologies while training universal GNN-based controllers via PPO. NGE demonstrates that GNN controllers can generalize across various body structures, thereby reducing the cost of controller retraining during morphology search. The scalability of ERL is further illustrated by \emph{Deep Evolutionary Reinforcement Learning} (DERL)~\cite{gupta2021embodied}. In contrast to NGE, DERL employs individual controllers that are trained from scratch for each morphology. Coupled with increasingly challenging environments, this framework enables the emergence of morphological intelligence, resulting in agents that exhibit increased robustness, adaptability, and behavioral diversity. Similar strategies appear in subsequent works~\cite{park2021computational,bhatia2021evolution,chen2023evolving,wang2024learn,yue2025toward,cheng2024structural,harada2024lamarckian,ringel2025text2robot,jin2025evolutionary}.

As summarized in Table~\ref{tab:bi-level}, recent research has expanded the classical ERL workflow along several dimensions. First, richer morphological representations have been introduced, including hyperbolic embeddings~\cite{dongleveraging2024}, text-to-mesh generation~\cite{ringel2025text2robot}, and latent space embedding~\cite{jeon2025convergent}. Second, advanced control-learning techniques, such as meta-RL~\cite{belmonte2022meta}, transfer learning~\cite{harada2024lamarckian}, imitation learning~\cite{hernandezfollowing}, and policy-sharing~\cite{xie2025accelerating,li2024generating,wang2024evolution}, have improved sample efficiency and generalization. A parallel line of work introduces novel morphological selection mechanisms, such as paired competition~\cite{banarse2019body}, successive halving~\cite{katayama2023ga}, multi-fidelity archiving~\cite{zhao2024morphological,nagiredla2024ecode}, and multi-objective balancing~\cite{zhao2025multi}, which can prevent premature convergence in morphology space~\cite{doncieux2014beyond,mertan2024investigating}.

Notably, the combination of EA and RL has also been widely explored in control optimization. These hybrid methods maintain a population of policies that evolve across generations, allowing for broad exploration of the policy space while refining individual controllers via RL’s gradient-based or value-based updates. Representative approaches include ERL~\cite{khadka2018evolution}, CERL~\cite{khadka2019collaborative}, PDERL~\cite{bodnar2020proximal}, SERL~\cite{wang2022surrogate}, and ERL-Re$^{2}$~\cite{hao2023erl}. Comprehensive reviews of this line of work can be found in~\cite{drugan2019reinforcement,sigaud2023combining,bai2023evolutionary,zhu2023survey,lin2025evolutionary,li2024bridging}.

\begin{table*}[htbp]
\renewcommand{\arraystretch}{0.85} 
\centering
\caption{Summary of bi-level co-design approaches.}
\vspace{-2mm}
\resizebox{\textwidth}{!}{
\begin{threeparttable} 
\begin{tabular}{c|l|c|c|c|ll|ll|ll}
\toprule
\multirow{2}[4]{*}{} & \multicolumn{1}{c|}{\multirow{2}[4]{*}{\textbf{Method}}} & \multicolumn{1}{c|}{\multirow{2}[4]{*}{\textbf{Year}}} & \multirow{2}[4]{*}{\textbf{\makecell{Agent\\Type}}} & \multirow{2}[4]{*}{\textbf{\makecell{Application\\(Sim/Real)}}} & \multicolumn{2}{c|}{\textbf{Controlling Brain}} & \multicolumn{2}{c|}{\textbf{Body Morphology}} & \multicolumn{2}{c}{\textbf{Task Environment}}\\
\cmidrule{6-11}      &       &       &       &       & \textbf{Design Space} & \textbf{Representation} & \textbf{Design Space} & \textbf{Representation}&\textbf{Design Space} & \textbf{Representation} \\
\midrule 
\multirow{4}[2]{*}{\rotatebox[origin=c]{90}{\textbf{Evolutionary Reinforcement Learning}}}
& \multirow{2}{*}{NGE~\cite{wang2019neural}} & \multirow{2}{*}{\makecell{2019\\~\href{https://www.cs.toronto.edu/~henryzhou/NGE_website/}{[link]}}} & \multirow{2}{*}{\makecell{Rigid\\Robot}} & \multirow{2}{*}{\makecell{Simulation\\\{MuJoCo\}}} & \cellcolor[rgb]{ .851,  .851,  .851}Learnable (CL) & \cellcolor[rgb]{ .851,  .851,  .851} Universal & \cellcolor[rgb]{ .851,  .851,  .851}Hybrid (UF) & \cellcolor[rgb]{ .851,  .851,  .851}Direct  & \cellcolor[rgb]{ .851,  .851,  .851}Predefined (Partial) & \cellcolor[rgb]{ .851,  .851,  .851}Task-Driven  \\
& & &  &  & RL (PPO) & GNN & EA (Mutation) & Graph & Locomotion& Task Reward\\

\cmidrule{2-11}
& \multirow{2}{*}{POEM~\cite{banarse2019body}} & \multirow{2}{*}{\makecell{2019\\~\href{https://www.youtube.com/watch?v=CHlecRim9PI}{[link]}}} & \multirow{2}{*}{\makecell{Rigid\\Robot}} & \multirow{2}{*}{\makecell{Simulation\\\{MuJoCo\}}} & \cellcolor[rgb]{ .851,  .851,  .851}Learnable (CL) & \cellcolor[rgb]{ .851,  .851,  .851} Individual & \cellcolor[rgb]{ .851,  .851,  .851}Continuous (F) & \cellcolor[rgb]{ .851,  .851,  .851}Direct  & \cellcolor[rgb]{ .851,  .851,  .851}Predefined (Full) & \cellcolor[rgb]{ .851,  .851,  .851}Task-Driven  \\
\bottomrule
\end{tabular}%
\end{threeparttable}       
} 
\label{tab:bi-level}
\vspace{-2mm}
\end{table*}   


\begin{table*}[htbp]
\renewcommand{\arraystretch}{0.85} 
\centering
\caption{Continued from previous page.}
\vspace{-2mm}
\resizebox{\textwidth}{!}{
\begin{threeparttable} 
\begin{tabular}{c|l|c|c|c|ll|ll|ll}
\toprule
\multirow{2}[4]{*}{} & \multicolumn{1}{c|}{\multirow{2}[4]{*}{\textbf{Method}}} & \multicolumn{1}{c|}{\multirow{2}[4]{*}{\textbf{Year}}} & \multirow{2}[4]{*}{\textbf{\makecell{Agent\\Type}}} & \multirow{2}[4]{*}{\textbf{\makecell{Application\\(Sim/Real)}}} & \multicolumn{2}{c|}{\textbf{Controlling Brain}} & \multicolumn{2}{c|}{\textbf{Body Morphology}} & \multicolumn{2}{c}{\textbf{Task Environment}}\\
\cmidrule{6-11}      &       &       &       &       & \textbf{Design Space} & \textbf{Representation} & \textbf{Design Space} & \textbf{Representation}&\textbf{Design Space} & \textbf{Representation} \\
\midrule

& \multirow{2}{*}{Bai et. al.~\cite{bai2025learning}} & \multirow{2}{*}{\makecell{2025\\~\href{}{[---]}}} & \multirow{2}{*}{\makecell{Soft \\ Manipulator}} & \multirow{2}{*}{\makecell{Sim\&Real \\\{Nvidia Warp\}}} & \cellcolor[rgb]{ .851,  .851,  .851}Predefined (CL) & \cellcolor[rgb]{ .851,  .851,  .851}Universal & \cellcolor[rgb]{ .851,  .851,  .851}Continuous (F) & \cellcolor[rgb]{ .851,  .851,  .851}Direct & \cellcolor[rgb]{ .851,  .851,  .851}Modifiable (Full)& \cellcolor[rgb]{ .851,  .851,  .851}Task-Driven\\
& & &  &  & None & Trajectories & EA (R-Model) & Vector & Manipulation & Task Reward \\

\midrule
& \multirow{2}{*}{Ha et al.~\cite{ha2018computational2}} & \multirow{2}{*}{\makecell{2018\\~\href{}{[---]}}} & \multirow{2}{*}{\makecell{Rigid\\ Robot}} & \multirow{2}{*}{\makecell{Sim\&Real \\\{DART\}}} & \cellcolor[rgb]{ .851,  .851,  .851}Learnable (OL) & \cellcolor[rgb]{ .851,  .851,  .851}Individual & \cellcolor[rgb]{ .851,  .851,  .851}Discrete (UF) & \cellcolor[rgb]{ .851,  .851,  .851}Direct & \cellcolor[rgb]{ .851,  .851,  .851}Predefined (Full)& \cellcolor[rgb]{ .851,  .851,  .851}Task-Driven\\
& & &  &  & Numerical Solver & Torques & Heuristic Search & Graph & Loco.\&Manip.& Task Reward\\
\bottomrule
\end{tabular}%
\begin{tablenotes}    
        \footnotesize               
        \item[] \textbf{Note}: The \textcolor{magenta}{[link]} in each row directs to the corresponding code or project website. The shaded section indicates the category of each item, while the white section below lists the specific methods and representations used. \emph{\textbf{CL}: Closed-Loop, \textbf{OL}: Open-Loop, \textbf{F}: Fixed, \textbf{UF}: Unfixed, \textbf{Loco.}: Locomotion, \textbf{Manip.}: Manipulation, \textbf{Bayesian-Opt}: Bayesian Optimization, \textbf{Grad. Opt.}: Gradient-Based Optimization, \textbf{Nonlinear-Opt}: Nonlinear Optimization, \textbf{DDP}: Differential Dynamic Programming, \textbf{LQR}: Linear Quadratic Regulators, \textbf{AME}: Asynchronous Morpho-Evolution,\textbf{Hom.}: Homeokinesis}
      \end{tablenotes}            
    \end{threeparttable}       
} 
\label{tab:bi-level2}
\vspace{-2mm}
\end{table*}

\subsubsection{Surrogate-Assisted Methods}
A central challenge in bi-level co-design arises from its hierarchical structure, where the inner-loop optimization is computationally expensive and often requires days of training for complex tasks. Surrogate-assisted methods address this issue by employing predictive models to approximate agent task performance, thereby reducing the need for repeated full-scale policy evaluations.

\emph{Bayesian Optimization} (BO)~\cite{shahriari2015taking} has emerged as a prominent framework for modeling the morphology–performance relationship in ECD. As a black-box, zeroth-order optimizer, BO efficiently handles expensive objective evaluations. At each iteration, BO updates a surrogate model using previously evaluated morphologies and their performance, then selects new candidates through acquisition functions, such as \emph{Probability of Improvement} (PI), \emph{Expected Improvement} (EI), \emph{Entropy Search} (ES), or \emph{Upper Confidence Bound} (UCB), thereby balancing exploration and exploitation~\cite{wilson2018maximizing}. The selected candidates are evaluated, and their results are used to update the surrogate model for the subsequent iteration.

\emph{Gaussian Processes} (GPs) are among the most commonly used surrogates in BO. For instance, HPC-BBO~\cite{liao2019data} introduced a hierarchical framework that facilitates parallel hardware validation. Similarly, \cite{bjelonic2023learning} pre-trained a universal controller in simulation and used BO to optimize parallel–elastic spring parameters for a real quadruped, demonstrating effective sim-to-real transfer. Variants of BO with GPs have been further explored in~\cite{bhatia2021evolution,hu2021neural,saito2024effective,pan2021emergent,rajani2023co}. To enhance generalization, \cite{chen2024pretraining} combined domain randomization with a pretraining–finetuning pipeline. Extending beyond single-task BO, \cite{dong2024cagecoopt} proposed a multi-task BO framework for robot manipulators. More recently, \cite{schneider2025task} combined BO with Hyperband to generate novel manipulator designs that improve performance across both seen and unseen tasks.

In addition to traditional surrogates used in BO, recent approaches have explored alternative choices. For example, Q-value functions (\emph{Q-Surro}) from RL have been used to evaluate candidate morphologies~\cite{luck2020data,luck2021robot,du2025efficient}. Multi-dynamics models can further enhance generalization across different agent designs~\cite{sliacka2023co}. Alternatively, when historical interaction data is available, near-optimal control actions (\emph{A-Surro}) can be partially or fully transferred from previously optimized designs to accelerate evaluation~\cite{liu2023rapidly}. More recent studies by~\cite{zhao2025cross,strgar2025accelerated} developed universal policies that span multiple tasks and can be efficiently employed as a zero-shot performance evaluator (\emph{A-Surro}), thereby improving efficiency. Moreover, in \cite{bai2025learning}, manipulator morphologies are evaluated and optimized using pre-collected trajectories and a learned reward model (\emph{R-Model}), with the reliance on reward model evaluations gradually increasing throughout the training process.

\subsubsection{Other Methods}
Notably, the inner-loop controller optimization can be implemented using alternatives to RL, including trajectory optimization~\cite{fadini2022simulation}, evolutionary algorithms~\cite{jelisavcic2019lamarckian,le2022morpho,le2025efficient}, numerical solvers~\cite{sartore2023codesign,ha2018computational2}, and information-theoretic classifiers~\cite{iii2021taskagnostic}. To accelerate bi-level co-design, Goff et al. proposed A-MEL~\cite{le2024improving}, which integrates asynchronous learning and evolution to reduce both time and computational costs. Strgar et al.~\cite{Strgar-RSS-24} further leveraged massively parallel differentiable simulations to efficiently optimize neural controllers for large morphology populations.

\subsection{Single-Level Co-Design}\label{single-level}
As shown in Fig.~\ref{fig_3} (b), single-level co-design treats both morphology and control as variables within a unified optimization process. We categorize recent work along this line into four subcategories: 1) \emph{\textbf{EA-Based Methods}}, 2) \emph{\textbf{RL-Based Methods}}, 3) \emph{\textbf{Physics-Based Methods}}, and 4) \emph{\textbf{Diversity-Based Methods}}, which are summarized in Table \ref{tab:single-level}.

\subsubsection{EA-Based Methods}
EAs have a long-standing history of jointly evolving robot morphology and control. In these methods, each \emph{individual} in the population encodes either a complete agent or only its morphology, along with a simplified controller. The evolutionary loop evaluates fitness and applies operators such as selection, mutation, and crossover to explore the combined design space.

Early work by Sims~\cite{sims1994evolving,sims1994competition} evolved 3D virtual creatures with embedded neural controllers, demonstrating the emergence of locomotion and showing that graph-based encodings naturally yield modular and symmetric morphologies. Subsequent research expanded this paradigm through diverse genotype representations, including developmental encodings~\cite{bongard2001repeated,brodbeck2015morphological}, CPPNs~\cite{auerbach2011evolving,tanaka2022co}, and learned morphological representations~\cite{zhao2025concurrent}. Recent efforts have introduced novel evolutionary search mechanisms. \emph{Morphological innovation protection}~\cite{cheney2018scalable} mitigates premature convergence by preserving newly evolved traits, while other studies incorporate heuristic search~\cite{ha2018computational}, biologically inspired strategies~\cite{davis2023subtract}, and neuroscience-inspired approaches~\cite{ferigo2025totipotent}. Moreover, multi-objective EAs, such as NSGA-II~\cite{deb2002fast}, have been utilized to balance competing design objectives and constraints. For example, minimizing energy consumption and mission time jointly in morphing aerial vehicles~\cite{bergonti2024co}.

EA-based methods excel at discovering novel morphologies due to their global and relatively unconstrained search capabilities. These methods have been widely used to generate complex geometries and behaviors~\cite{cheney2013unshackling,hiller2010evolving,cheney2014electro,hiller2012automatic,kriegman2021fractals,sathuluri2023robust}. However, scalability remains a major limitation: population-based evaluation is computationally expensive—often requiring thousands of simulations per generation—and convergence is typically slow, mirroring the efficiency bottlenecks observed in bi-level co-design.

\subsubsection{RL-Based Methods}\label{rlb}
RL has proven effective for solving sequential decision-making problems, making it a natural tool for single-level embodied co-design. Modern RL-based approaches often consider control and morphology as jointly learnable parameters or approximate gradients for both.

Early methods incorporated morphology directly into the policy representation. For example, \cite{ha2019reinforcement} used a population-based policy-gradient method (REINFORCE) to update both control and morphology simultaneously. Similarly, \cite{Chen2020HardwareAP} embedded body parameters into the computational graph of a RL policy. \cite{schaff2019jointly} co-optimized a differentiable distribution over morphology parameters alongside a universal controller; this framework was later extended by \cite{schaff2022n} for more complex embodiments. Other works, such as \cite{whitman2020modular2}, employed Q-learning to guide modular manipulator design.

A complementary paradigm treats design modifications as actions within the RL framework. Under this formulation, the agent concurrently learns to control and perform morphological ``design actions”. A representative work is Transform2Act~\cite{yuan2022transformact}, which introduces a transform-and-control paradigm that formulates design optimization as a conditional policy over both control and morphology. This approach incorporates policy-gradient updates and experience sharing across designs to improve sample efficiency. Subsequent extensions have incorporated curriculum learning~\cite{wang2023curriculum}, multi-task pretraining~\cite{wang2023preco}, symmetry priors~\cite{dong2023symmetry}, hierarchical modeling~\cite{sun2023co}, competition mechanisms~\cite{huang2024competevo}, and temporal credit assignment~\cite{lu2025bodygen}, with applications ranging from embodied manipulation~\cite{liu2023learning,kulz2025design} to freeform modular robot design~\cite{li2024reinforcement}. Moreover, real-world considerations have motivated sim-to-real approaches. \cite{schaff2022soft} showed that co-designed soft robots can effectively navigate complex terrains. MORPH~\cite{he2024morph} co-optimized a neural proxy of hardware alongside the control policy, enabling gradient-based updates even with non-differentiable physics. 

Notably, another related direction investigates the real-time adaptation of morphology and control using RL. Pathak et al.~\cite{pathak2019learning} demonstrated that modular, self-assembling agents can learn to reconfigure their bodies online, with individual modules attaching or detaching to form composite entities that enhance adaptability across diverse tasks. Similar approaches have been applied to soft robots with dynamically deformable morphologies~\cite{huang2024dittogym,pigozzi2024pressure}, rigid robots with reconfigurable structures~\cite{chiappa2022dmap}, and block-based robotic assemblies~\cite{ghasemipour2022blocks}.

\subsubsection{Physics-Based Methods}\label{phy}
Another approach to single-level co-design leverages physics-based optimization. If the simulator, analytic model, or learned dynamics is differentiable, gradients can be computed with respect to both the controller and body parameters. Task losses can then be backpropagated through time using gradient-based optimizers (\emph{Grad. Opt.}), such as ADAM~\cite{adam} or L-BFGS~\cite{zhu1997algorithm}. Compared with black-box EA search or gradient approximations in RL, these approaches exhibit greater sample efficiency.

Early efforts relied on parametric analytic models~\cite{spielberg2017functional,ha2018computational} and adjoint methods~\cite{desai2018interactive} to model the complex relationships between morphology and control parameters. Subsequent research introduced stochastic programming in combination with trajectory optimization~\cite{bravo2020one,bravo2024engineering}, thereby improving scalability across tasks and environments. More recent methods incorporate hardware and environmental constraints~\cite{ghansah2023humanoid,vaish2024co,sartore2022optimization}. Most analytic approaches assume fixed body topologies, primarily due to modeling challenges. Differentiable simulators address this limitation by providing end-to-end gradients through various physical processes, such as soft-body~\cite{du2021diffpd,ma2021diffaqua,qiao2021differentiable,hu2019chainqueen,hu2019difftaichi}, rigid-body~\cite{freeman2021brax,xu2021end,degrave2019differentiable}, and differentiable rendering~\cite{murthy2020gradsim,ma2021risp}. These processes enable gradients to propagate through more expressive design spaces. Built on this foundation, deformation-based~\cite{xu2021end,cai2023task2morph,liu2023learning,lirobottool2023} and density-based schemes~\cite{yuhn20234d,matthews2023efficient,sato2023topology,sato2025computational} have been proposed for agents with complex geometries. Additional mechanisms include particle sets~\cite{spielberg2021co,matthews2023efficient}, Wasserstein barycenters~\cite{ma2021diffaqua,wang2023softzoo}, and CPPN-based encodings~\cite{cochevelou2023differentiable}.

Physics-based co-design can scale to very high-dimensional bodies (e.g., thousands of masses or ``voxels”) because a single gradient pass provides information about all design parameters simultaneously. However, accurate differentiable models for contact, fluid dynamics, and soft-material deformation remain limited. These models often require smooth design spaces, along with simplified controllers (e.g., open-loop torque control), resulting in relatively simple behaviors. Combining model-based gradients with model-free RL is an emerging trend to overcome these limitations~\cite{spielberg2019learning}.

\begin{table*}[htbp]
\renewcommand{\arraystretch}{0.85} 
\centering
\caption{Summary of Single-Level Co-Design approaches.}
\vspace{-2mm}
\resizebox{\textwidth}{!}{
\begin{threeparttable} 
\begin{tabular}{c|l|c|c|c|ll|ll|ll}
\toprule
\multirow{2}[4]{*}{} & \multicolumn{1}{c|}{\multirow{2}[4]{*}{\textbf{Method}}} & \multicolumn{1}{c|}{\multirow{2}[4]{*}{\textbf{Year}}} & \multirow{2}[4]{*}{\textbf{\makecell{Agent\\Type}}} & \multirow{2}[4]{*}{\textbf{\makecell{Application\\(Sim/Real)}}} & \multicolumn{2}{c|}{\textbf{Controlling Brain}} & \multicolumn{2}{c|}{\textbf{Body Morphology}} & \multicolumn{2}{c}{\textbf{Task Environment}}\\
\cmidrule{6-11}      &       &       &       &       & \textbf{Design Space} & \textbf{Representation} & \textbf{Design Space} & \textbf{Representation}&\textbf{Design Space} & \textbf{Representation} \\
\midrule 
\multirow{4}[2]{*}{\rotatebox[origin=c]{90}{\textbf{EA-Based Methods}}}
& \multirow{2}{*}{Auerbach et al.~\cite{auerbach2011evolving}} & \multirow{2}{*}{\makecell{2011\\~\href{}{[---]}}} & \multirow{2}{*}{\makecell{Modular\\Robot}} & \multirow{2}{*}{\makecell{Simulation\\\{ODE\}}} & \cellcolor[rgb]{ .851,  .851,  .851}Learnable (CL) & \cellcolor[rgb]{ .851,  .851,  .851}Individual & \cellcolor[rgb]{ .851,  .851,  .851}Continuous (UF) & \cellcolor[rgb]{ .851,  .851,  .851}Indirect & \cellcolor[rgb]{ .851,  .851,  .851}Predefined (Full)& \cellcolor[rgb]{ .851,  .851,  .851}Task‑Driven\\
& & &  &  & CPPN-NEAT & RNN  & CPPN-NEAT & CPPN & Navigation& Task Reward\\

\cmidrule{2-11}
& \multirow{2}{*}{Hiller et al.~\cite{hiller2012automatic}} & \multirow{2}{*}{\makecell{2012\\~\href{}{[---]}}} & \multirow{2}{*}{\makecell{Soft\\Robot}} & \multirow{2}{*}{\makecell{Sim\&Real\\\{Customized\}}} & \cellcolor[rgb]{ .851,  .851,  .851}Predefined (OL) & \cellcolor[rgb]{ .851,  .851,  .851}Universal & \cellcolor[rgb]{ .851,  .851,  .851}Continuous (UF) & \cellcolor[rgb]{ .851,  .851,  .851}Direct & \cellcolor[rgb]{ .851,  .851,  .851}Predefined (Full)& \cellcolor[rgb]{ .851,  .851,  .851}Task-Driven\\
& & &  &  & None & CPG & EA & GMM & Locomotion&Task Reward \\
\bottomrule
\end{tabular}%
\end{threeparttable}       
} 
\label{tab:single-level}
\vspace{-2mm}
\end{table*}   


\begin{table*}[htbp]
\renewcommand{\arraystretch}{0.85} 
\centering
\caption{Continued from previous page.}
\vspace{-2mm}
\resizebox{\textwidth}{!}{
\begin{threeparttable} 
\begin{tabular}{c|l|c|c|c|ll|ll|ll}
\toprule
\multirow{2}[4]{*}{} & \multicolumn{1}{c|}{\multirow{2}[4]{*}{\textbf{Method}}} & \multicolumn{1}{c|}{\multirow{2}[4]{*}{\textbf{Year}}} & \multirow{2}[4]{*}{\textbf{\makecell{Agent\\Type}}} & \multirow{2}[4]{*}{\textbf{\makecell{Application\\(Sim/Real)}}} & \multicolumn{2}{c|}{\textbf{Controlling Brain}} & \multicolumn{2}{c|}{\textbf{Body Morphology}} & \multicolumn{2}{c}{\textbf{Task Environment}}\\
\cmidrule{6-11}      &       &       &       &       & \textbf{Design Space} & \textbf{Representation} & \textbf{Design Space} & \textbf{Representation}&\textbf{Design Space} & \textbf{Representation} \\
\midrule

& \multirow{2}{*}{Cochevelou et al.~\cite{cochevelou2023differentiable}} & \multirow{2}{*}{\makecell{2023\\~\href{}{[---]}}} & \multirow{2}{*}{\makecell{Soft \\Robot}} & \multirow{2}{*}{\makecell{Simulation \\\{Taichi\}}} & \cellcolor[rgb]{ .851,  .851,  .851}Learnable (OL) & \cellcolor[rgb]{ .851,  .851,  .851}Individual & \cellcolor[rgb]{ .851,  .851,  .851}Continuous (UF) & \cellcolor[rgb]{ .851,  .851,  .851}Direct & \cellcolor[rgb]{ .851,  .851,  .851}Predefined (Full)& \cellcolor[rgb]{ .851,  .851,  .851}Task-Driven\\
& & &  &  & Grad. Opt. & CPG & Grad. Opt. & Tensor & Locomotion & Task Loss \\

\cmidrule{2-11}
& \multirow{2}{*}{Ghansah et al.~\cite{ghansah2023humanoid}} & \multirow{2}{*}{\makecell{2023\\~\href{}{[---]}}} & \multirow{2}{*}{\makecell{Humanoid \\Robot}} & \multirow{2}{*}{\makecell{Sim\&Real \\\{Customized\}}} & \cellcolor[rgb]{ .851,  .851,  .851}Learnable (CL) & \cellcolor[rgb]{ .851,  .851,  .851}Individual & \cellcolor[rgb]{ .851,  .851,  .851}Continuous (F) & \cellcolor[rgb]{ .851,  .851,  .851}Direct & \cellcolor[rgb]{ .851,  .851,  .851}Predefined (Full)& \cellcolor[rgb]{ .851,  .851,  .851}Task-Driven\\
& & &  &  & Traj. Opt. & Torques & Traj. Opt. & Vector & Locomotion & Task Loss \\
\bottomrule
\end{tabular}%
\begin{tablenotes}    
        \footnotesize               
        \item[] \textbf{Note}: The \textcolor{magenta}{[link]} in each row directs to the corresponding code or project website. The shaded section indicates the category of each item, while the white section below lists the specific methods and representations used. For RL-based methods, we further specify their design policy representations. \emph{\textbf{CL}: Closed-Loop, \textbf{OL}: Open-Loop, \textbf{F}: Fixed, \textbf{UF}: Unfixed, \textbf{Loco.}: Locomotion, \textbf{Manip.}: Manipulation, \textbf{ID}: Inverse
 Dynamics, \textbf{GMM}: Gaussian Mixture Model, \textbf{Policy Grad.}: Policy Gradient, \textbf{Dist. Adapt.}: Distribution Adaptation, \textbf{DQN}: Deep Q Network, \textbf{Grad. Opt.}: Gradient-Based Optimization, \textbf{Trans.}: Transformer, \textbf{Gibbs.}: Gibbs Distribution, \textbf{Wass. Bary.}: Wasserstein Barycenter, \textbf{Stoc. Prog.}: Stochastic
 Programming, \textbf{Traj. Opt.}: Trajectory Optimization, \textbf{NLP}:Nonlinear Programming, \textbf{ANN}: Artificial Neural Network }
      \end{tablenotes}            
\end{threeparttable}       
} 
\label{tab:single-level2}
\vspace{-2mm}
\end{table*}

\subsubsection{Diversity-Based Methods}\label{diver}
A persistent challenge in single-level ECD is the premature convergence of both morphology and controller. Diversity-based methods address this issue by optimizing a \emph{repertoire} of high-quality body–brain pairs, rather than concentrating on a single global optimum.

Early work by \cite{krvcah2010solving,lehman2011evolving} demonstrated this principle by employing \emph{Novelty Search}~\cite{doncieux2019novelty} with local competition in a three-dimensional morphology space (height, mass, and joint count), showing that localized selection pressures help preserve morphological diversity. The subsequent work built upon \emph{Quality-Diversity} (QD) algorithms~\cite{pugh2016quality,cully2017quality}. Canonical QD methods target only behavioral diversity. For example, MAP-Elites~\cite{mouret2015illuminating} discretizes a user-defined \emph{behavior descriptor} (BD) space into cells and stores the best-performing solution within each cell. In the context of ECD, \emph{morphology descriptors} (MDs) are introduced to index MAP-Elites archives. For instance, \cite{nordmoen2020quality,nordmoen2021map} discretized a two-dimensional MD space (total voxels and active voxels) and stored the fastest walker in each cell, achieving greater robustness and diversity compared to objective-only EAs. Recent works extend this framework to legged robots and manipulators~\cite{howard2020diversity,xie2024map,norstein2023effects}, or distill the archive into a universal controller to stabilize the co-design process~\cite{mertan2025controller}. Moreover, dual-map approaches separately represent morphology and controller spaces~\cite{zardini2021seeking}, while 3B-QD~\cite{nadizar2025enhancing} maintains three independent archives for morphology, control, and behavior, further enhancing the illumination of the agent co-design space.

Despite their effectiveness, diversity-based ECD faces several challenges, including the automatic discovery of informative descriptors, scalability to high-dimensional material and property spaces~\cite{pigozzi2023factors,mertan2024no}, and the lack of theoretical guarantees for niche-optimal solutions. Comparative evaluations of QD-based approaches in ECD can be found in~\cite{mkhatshwa2023impact,mkhatshwa2025body}.

\subsection{Generative Co-Design}\label{genertive}
Generative co-design leverages automated generative models to capture the relationships among morphology, tasks, and performance (evaluated by learning a controller). This approach enables optimization within latent or rule-defined spaces, as illustrated in Fig.~\ref{fig_3} (c). We further categorize the generative co-design framework into three sub-topics: 1) \emph{\textbf{Rule-Based Generation}}, 2) \emph{\textbf{Latent-Based Generation}}, and 3) \emph{\textbf{Large-Model-Based Generation}}. Each class employs a distinct generative representation and search strategy to explore morphology spaces, as summarized in Table~\ref{tab:gen}.

\subsubsection{Rule-Based Generation}
Early work along this line typically utilized graph-based generative encodings~\cite{sims1994competition,sims1994evolving,pilat2008creature} and L-systems~\cite{hornby2001evolution,hornby2001evolving,hornby2003generative} to jointly generate morphology and controller parameters, which were then evolved to create task-oriented agents. More recent approaches combine grammars with advanced search strategies. Robogrammar~\cite{zhao2020robogrammar} and its extension~\cite{xu2021multi} encode robots as graphs, enforce valid assemblies through a graph grammar, and couple heuristic search with a model-based controller to efficiently identify high-performing designs. Zhao et al.~\cite{zhao2022automatic} extended this framework to aerial vehicles by combining discrete grammar exploration with a differentiable simulator, enabling both rediscovery of canonical drones as well as the discovery of novel configurations. Zharkov et al.~\cite{zharkov2024synergizing} further applied grammar-based linkage graphs to co-design underactuated, tendon-driven grippers.

Rule-based approaches effectively constrain the search space to plausible designs, explicitly encode prior knowledge (e.g., symmetry and connectivity), and improve manufacturability. However, their reliance on expert-crafted grammars can introduce bias and restrict the exploration of novel or fine-grained morphological variations.

\begin{table*}[htbp]
\renewcommand{\arraystretch}{0.9} 
\centering
\caption{Summary of Generative Co-Design approaches.}
\vspace{-2mm}
\resizebox{\textwidth}{!}{
\begin{threeparttable} 
\begin{tabular}{c|l|c|c|c|ll|ll|ll}
\toprule
\multirow{2}[4]{*}{} & \multicolumn{1}{c|}{\multirow{2}[4]{*}{\textbf{Method}}} & \multicolumn{1}{c|}{\multirow{2}[4]{*}{\textbf{Year}}} & \multirow{2}[4]{*}{\textbf{\makecell{Agent\\Type}}} & \multirow{2}[4]{*}{\textbf{\makecell{Application\\(Sim/Real)}}} & \multicolumn{2}{c|}{\textbf{Controlling Brain}} & \multicolumn{2}{c|}{\textbf{Body Morphology}} & \multicolumn{2}{c}{\textbf{Task Environment}}\\
\cmidrule{6-11}      &       &       &       &       & \textbf{Design Space} & \textbf{Representation} & \textbf{Design Space} & \textbf{Representation}&\textbf{Design Space} & \textbf{Representation} \\
\midrule 
\multirow{4}[2]{*}{\rotatebox[origin=c]{90}{\textbf{Rule-Based}}} 
& \multirow{2}{*}{RoboGrammar~\cite{zhao2020robogrammar}} & \multirow{2}{*}{\makecell{2020\\~\href{https://github.com/allanzhao/RoboGrammar}{[link]}}} & \multirow{2}{*}{\makecell{Rigid\\Robot}} & \multirow{2}{*}{\makecell{Simulation\\\{Bullet\}}} & \cellcolor[rgb]{ .851,  .851,  .851}Learnable (CL) & \cellcolor[rgb]{ .851,  .851,  .851}Individual & \cellcolor[rgb]{ .851,  .851,  .851}Discrete (UF) & \cellcolor[rgb]{ .851,  .851,  .851}Indirect & \cellcolor[rgb]{ .851,  .851,  .851}Predefined (Full)& \cellcolor[rgb]{ .851,  .851,  .851}Task‑Driven\\
& & &  &  & MPPI & MPC & GHS & Grammar & Locomotion& Task Reward\\

 \cmidrule{2-11}
& \multirow{2}{*}{MOGHS~\cite{xu2021multi}} & \multirow{2}{*}{\makecell{2021\\~\href{https://moghs.csail.mit.edu/}{[link]}}} & \multirow{2}{*}{\makecell{Rigid\\Robot}} & \multirow{2}{*}{\makecell{Simulation\\\{Bullet\}}} & \cellcolor[rgb]{ .851,  .851,  .851}Learnable (CL) & \cellcolor[rgb]{ .851,  .851,  .851}Individual & \cellcolor[rgb]{ .851,  .851,  .851}Discrete (UF) & \cellcolor[rgb]{ .851,  .851,  .851}Indirect & \cellcolor[rgb]{ .851,  .851,  .851}Predefined (Full)& \cellcolor[rgb]{ .851,  .851,  .851}Task‑Driven\\
& & &  &  & MPPI & MPC & GHS & Grammar & Locomotion& Task Reward\\

\cmidrule{2-11}
& \multirow{2}{*}{Zhao et al.~\cite{zhao2022automatic}} & \multirow{2}{*}{\makecell{2022\\~\href{}{[---]}}} & \multirow{2}{*}{\makecell{Aerial\\Vehicle}} & \multirow{2}{*}{\makecell{Simulation\\\{Customized\}}} & \cellcolor[rgb]{ .851,  .851,  .851}Learnable (CL) & \cellcolor[rgb]{ .851,  .851,  .851}Individual & \cellcolor[rgb]{ .851,  .851,  .851}Hybrid (UF) & \cellcolor[rgb]{ .851,  .851,  .851}Hybrid & \cellcolor[rgb]{ .851,  .851,  .851}Predefined (Full)& \cellcolor[rgb]{ .851,  .851,  .851}Task‑Driven\\
& & &  &  & Grad. Opt. & LQR & Grad. Opt.\& GHS & Grammar\&Vector & Flight& Task Reward\\

\cmidrule{2-11}
& \multirow{2}{*}{Zharkov et al.~\cite{zharkov2024synergizing}} & \multirow{2}{*}{\makecell{2024\\~\href{https://github.com/aimclub/rostok}{[link]}}} & \multirow{2}{*}{\makecell{Robot\\Manipulator}} & \multirow{2}{*}{\makecell{Sim\&Real\\\{Rostok\}}} & \cellcolor[rgb]{ .851,  .851,  .851}Predefined (OL) & \cellcolor[rgb]{ .851,  .851,  .851}Individual & \cellcolor[rgb]{ .851,  .851,  .851}Discrete (UF) & \cellcolor[rgb]{ .851,  .851,  .851}Indirect  & \cellcolor[rgb]{ .851,  .851,  .851}Predefined (Full)& \cellcolor[rgb]{ .851,  .851,  .851}Task‑Driven\\
& & &  &  & None & Torques & MCTS & Grammar & Manipulation& Task Reward\\

\midrule
\multirow{3}[2]{*}{\rotatebox[origin=c]{90}{\textbf{Latent-Based}}} 
& \multirow{2}{*}{Pan et al.~\cite{pan2021emergent}} & \multirow{2}{*}{\makecell{2021\\~\href{ https://xinleipan.github.io/emergent_morphology/}{[link]}}} & \multirow{2}{*}{\makecell{Robot\\Manipulator}} & \multirow{2}{*}{\makecell{Simulation\\\{Bullet\}}} & \cellcolor[rgb]{ .851,  .851,  .851}Learnable (OL) & \cellcolor[rgb]{ .851,  .851,  .851}Individual & \cellcolor[rgb]{ .851,  .851,  .851}Hybrid (UF) & \cellcolor[rgb]{ .851,  .851,  .851}Indirect  & \cellcolor[rgb]{ .851,  .851,  .851}Modifiable (Full)& \cellcolor[rgb]{ .851,  .851,  .851}Hybrid\\
& & &  &  & Bayesian-Opt & Torques & Bayesian-Opt & Latent (VAE) & Manipulation& Task+Complexity \\

\cmidrule{2-11}
& \multirow{2}{*}{RoboGAN~\cite{hu2022modular}} & \multirow{2}{*}{\makecell{2022\\~\href{}{[---]}}} & \multirow{2}{*}{\makecell{Modular\\Robot}} & \multirow{2}{*}{\makecell{Simulation\\\{Bullet\}}} & \cellcolor[rgb]{ .851,  .851,  .851}Predefined (CL) & \cellcolor[rgb]{ .851,  .851,  .851}Universal & \cellcolor[rgb]{ .851,  .851,  .851}Discrete (F) & \cellcolor[rgb]{ .851,  .851,  .851}Indirect & \cellcolor[rgb]{ .851,  .851,  .851}Modifiable (Full)& \cellcolor[rgb]{ .851,  .851,  .851}Task‑Driven\\
& & &  &  & RL & DNN & EA (ES) & Latent (GAN) & Locomotion& Task Reward\\

\cmidrule{2-11}
& \multirow{2}{*}{GLSO~\cite{hu2023glso}} & \multirow{2}{*}{\makecell{2023\\~\href{https://github.com/JiahengHu/GLSO}{[link]}}} & \multirow{2}{*}{\makecell{Modular\\Robot}} & \multirow{2}{*}{\makecell{Simulation\\\{Bullet\}}} & \cellcolor[rgb]{ .851,  .851,  .851}Learnable (CL) & \cellcolor[rgb]{ .851,  .851,  .851}Individual & \cellcolor[rgb]{ .851,  .851,  .851}Discrete (UF) & \cellcolor[rgb]{ .851,  .851,  .851}Indirect & \cellcolor[rgb]{ .851,  .851,  .851}Predefined (Full)& \cellcolor[rgb]{ .851,  .851,  .851}Task‑Driven\\
& & &  &  & MPPI & MPC & Bayesian-Opt & Graph (VAE) & Locomotion& Task Reward\\

\cmidrule{2-11}
& \multirow{2}{*}{Diffusebot~\cite{wang2023diffusebot}} & \multirow{2}{*}{\makecell{2023\\~\href{https://diffusebot.github.io/}{[link]}}} & \multirow{2}{*}{\makecell{Soft\\Robot}} & \multirow{2}{*}{\makecell{Sim\&Real\\\{SoftZoo\}}} & \cellcolor[rgb]{ .851,  .851,  .851}Learnable(OL) & \cellcolor[rgb]{ .851,  .851,  .851}Individual & \cellcolor[rgb]{ .851,  .851,  .851}Continuous (UF) & \cellcolor[rgb]{ .851,  .851,  .851}Indirect & \cellcolor[rgb]{ .851,  .851,  .851}Predefined (Full)& \cellcolor[rgb]{ .851,  .851,  .851}Task-Driven\\
& & &  &  & Grad. Opt. & CPG & Grad. Opt. & Latent (Diffusion) &Loco.\&Manip.& Task Reward\\

\cmidrule{2-11}
& \multirow{2}{*}{MorphVAE~\cite{song2024morphvae}} & \multirow{2}{*}{\makecell{2024\\~\href{https://github.com/WoodySJR/MorphVAE}{[link]}}} & \multirow{2}{*}{\makecell{Modular\\ Soft Robot}} & \multirow{2}{*}{\makecell{Simulation\\\{EvoGym\}}} & \cellcolor[rgb]{ .851,  .851,  .851}Learnable (CL) & \cellcolor[rgb]{ .851,  .851,  .851}Individual & \cellcolor[rgb]{ .851,  .851,  .851}Discrete (UF) & \cellcolor[rgb]{ .851,  .851,  .851}Indirect & \cellcolor[rgb]{ .851,  .851,  .851}Predefined (Full)& \cellcolor[rgb]{ .851,  .851,  .851}Task‑Driven\\
& & &  &  & RL (PPO) & DNN & Selection & Latent (VAE) & Loco.\&Manip. & Task Reward \\

\cmidrule{2-11}
& \multirow{2}{*}{RoboNet~\cite{nagiredla2024robonet}} & \multirow{2}{*}{\makecell{2024\\~\href{}{[---]}}} & \multirow{2}{*}{\makecell{Modular\\ Robot}} & \multirow{2}{*}{\makecell{Simulation\\\{MuJoCo\}}} & \cellcolor[rgb]{ .851,  .851,  .851}Learnable (CL) & \cellcolor[rgb]{ .851,  .851,  .851}Individual & \cellcolor[rgb]{ .851,  .851,  .851}Discrete (UF) & \cellcolor[rgb]{ .851,  .851,  .851}Indirect & \cellcolor[rgb]{ .851,  .851,  .851}Predefined (Full)& \cellcolor[rgb]{ .851,  .851,  .851}Task-Driven\\
& & &  &  & RL & DNN & Grad. Opt. & Graph (GFlowNet) & Locomotion & Task Reward\\

\cmidrule{2-11}
& \multirow{2}{*}{DGDM~\cite{xu2024dynamics}} & \multirow{2}{*}{\makecell{2024\\~\href{https://dgdmcorl.github.io/}{[link]}}} & \multirow{2}{*}{\makecell{Robot\\Gripper}} & \multirow{2}{*}{\makecell{Sim\&Real\\\{MuJoCo\}}} & \cellcolor[rgb]{ .851,  .851,  .851}Predefined (OL) & \cellcolor[rgb]{ .851,  .851,  .851}Universal & \cellcolor[rgb]{ .851,  .851,  .851}Continuous (UF) & \cellcolor[rgb]{ .851,  .851,  .851}Indirect & \cellcolor[rgb]{ .851,  .851,  .851}Modifiable (Full)& \cellcolor[rgb]{ .851,  .851,  .851}Task-Driven\\
& & &  &  & None & Torques & Grad. Opt. & Latent (Diffusion) & Manipulation& Task Reward\\

\cmidrule{2-11}
& \multirow{2}{*}{TERoboNet~\cite{nagiredlate}} & \multirow{2}{*}{\makecell{2025\\~\href{}{[---]}}} & \multirow{2}{*}{\makecell{Modular\\ Robot}} & \multirow{2}{*}{\makecell{Simulation\\\{MuJoCo\}}} & \cellcolor[rgb]{ .851,  .851,  .851}Learnable (CL) & \cellcolor[rgb]{ .851,  .851,  .851}Hybrid & \cellcolor[rgb]{ .851,  .851,  .851}Discrete (UF) & \cellcolor[rgb]{ .851,  .851,  .851}Indirect & \cellcolor[rgb]{ .851,  .851,  .851}Predefined (Full)& \cellcolor[rgb]{ .851,  .851,  .851}Task-Driven\\
& & &  &  & RL (PPO) & DNN & Grad. Opt. & Graph (GFlowNet) & Locomotion & Task Reward\\

\cmidrule{2-11}
& \multirow{2}{*}{COGNet~\cite{nagiredlacogent}} & \multirow{2}{*}{\makecell{2025\\~\href{}{[---]}}} & \multirow{2}{*}{\makecell{Modular\\ Robot}} & \multirow{2}{*}{\makecell{Simulation\\\{MuJo\&EvoG\}}} & \cellcolor[rgb]{ .851,  .851,  .851}Learnable (CL) & \cellcolor[rgb]{ .851,  .851,  .851}Individual & \cellcolor[rgb]{ .851,  .851,  .851}Discrete (UF) & \cellcolor[rgb]{ .851,  .851,  .851}Indirect & \cellcolor[rgb]{ .851,  .851,  .851}Predefined (Full)& \cellcolor[rgb]{ .851,  .851,  .851}Task-Driven\\
& & &  &  & RL (PPO) & DNN & Grad. Opt. & Graph (GFlowNet) & Locomotion & Task Reward\\

\cmidrule{2-11}
& \multirow{2}{*}{TAMS~\cite{liu2025terrain}} & \multirow{2}{*}{\makecell{2025\\~\href{https://github.com/TypeFloat/TAMS}{[link]}}} & \multirow{2}{*}{\makecell{Modular\\ Robot}} & \multirow{2}{*}{\makecell{Simulation\\\{MuJoCo\}}} & \cellcolor[rgb]{ .851,  .851,  .851}Learnable (CL) & \cellcolor[rgb]{ .851,  .851,  .851}Individual & \cellcolor[rgb]{ .851,  .851,  .851}Discrete (UF) & \cellcolor[rgb]{ .851,  .851,  .851}Indirect & \cellcolor[rgb]{ .851,  .851,  .851}Modifiable (Full)& \cellcolor[rgb]{ .851,  .851,  .851}Task-Driven\\
& & &  &  & MPPI & MPC & EA (DE) & Latent (VAE) & Locomotion & Task Reward\\

\midrule
\multirow{3}[2]{*}{\rotatebox[origin=c]{90}{\textbf{Large-Model-Based}}} 
& \multirow{2}{*}{OpenELM~\cite{lehman2023evolution}} & \multirow{2}{*}{\makecell{2025\\~\href{https://github.com/CarperAI/OpenELM}{[link]}}} & \multirow{2}{*}{\makecell{Sodarace \\ Robot}} & \multirow{2}{*}{\makecell{Simulation\\\{Sodarace\}}} & \cellcolor[rgb]{ .851,  .851,  .851}Learnable (OL) & \cellcolor[rgb]{ .851,  .851,  .851}Individual & \cellcolor[rgb]{ .851,  .851,  .851}Hybrid (UF) & \cellcolor[rgb]{ .851,  .851,  .851}Indirect & \cellcolor[rgb]{ .851,  .851,  .851}Modifiable (Full)& \cellcolor[rgb]{ .851,  .851,  .851}Hybrid \\
& & &  &  & LLM with GP & CPG & LLM with GP & Python Program & Locomotion & Task+Diversity\\


\cmidrule{2-11}
& \multirow{2}{*}{LASeR~\cite{song2025laser}} & \multirow{2}{*}{\makecell{2025\\~\href{https://github.com/WoodySJR/LASeR}{[link]}}} & \multirow{2}{*}{\makecell{Modular\\ Soft Robot}} & \multirow{2}{*}{\makecell{Simulation\\\{EvoGym\}}} & \cellcolor[rgb]{ .851,  .851,  .851}Learnable (CL) & \cellcolor[rgb]{ .851,  .851,  .851}Individual & \cellcolor[rgb]{ .851,  .851,  .851}Discrete (UF) & \cellcolor[rgb]{ .851,  .851,  .851}Direct & \cellcolor[rgb]{ .851,  .851,  .851}Predefined (Full)& \cellcolor[rgb]{ .851,  .851,  .851}Hybrid\\
& & &  &  & RL (PPO) & DNN & EA (LLM) & Matrix & Loco.\&Manip. & Task+Diversity\\

\cmidrule{2-11}
& \multirow{2}{*}{RoboMoRe~\cite{fang2025robomore}} & \multirow{2}{*}{\makecell{2025\\~\href{https://github.com/morphing-matter-lab/RoboMoRe}{[link]}}} & \multirow{2}{*}{\makecell{Rigid\\ Robot}} & \multirow{2}{*}{\makecell{Simulation\\\{MuJoCo\}}} & \cellcolor[rgb]{ .851,  .851,  .851}Learnable (CL) & \cellcolor[rgb]{ .851,  .851,  .851}Individual & \cellcolor[rgb]{ .851,  .851,  .851}Continuous (F) & \cellcolor[rgb]{ .851,  .851,  .851}Direct & \cellcolor[rgb]{ .851,  .851,  .851}Modifiable (Full)& \cellcolor[rgb]{ .851,  .851,  .851}Task-Driven\\
& & &  &  & RL (SAC) & DNN & LLM & XML & Locomotion & LLM-Generated\\

\cmidrule{2-11}
& \multirow{2}{*}{RoboMorph~\cite{qiu2024robomorph}} & \multirow{2}{*}{\makecell{2025\\~\href{}{[---]}}} & \multirow{2}{*}{\makecell{Modular \\ Robot}} & \multirow{2}{*}{\makecell{Simulation\\\{MuJoCo\}}} & \cellcolor[rgb]{ .851,  .851,  .851}Learnable (CL) & \cellcolor[rgb]{ .851,  .851,  .851}Individual & \cellcolor[rgb]{ .851,  .851,  .851}Discrete (UF) & \cellcolor[rgb]{ .851,  .851,  .851}Indirect & \cellcolor[rgb]{ .851,  .851,  .851}Predefined (Full)& \cellcolor[rgb]{ .851,  .851,  .851}Task-Driven\\
& & &  &  & RL (SAC) & DNN & EA (LLM) & Grammar & Locomotion & Task Reward\\

\cmidrule{2-11}
& \multirow{2}{*}{VLMGINEER~\cite{gao2025vlmgineer}} & \multirow{2}{*}{\makecell{2025\\~\href{https://vlmgineer.github.io/}{[link]}}} & \multirow{2}{*}{\makecell{End Effector
\\Tools}} & \multirow{2}{*}{\makecell{Simulation\\\{Bullet\}}} & \cellcolor[rgb]{ .851,  .851,  .851}Learnable (OL) & \cellcolor[rgb]{ .851,  .851,  .851}Individual & \cellcolor[rgb]{ .851,  .851,  .851}Continuous (UF) & \cellcolor[rgb]{ .851,  .851,  .851}Direct & \cellcolor[rgb]{ .851,  .851,  .851}Predefined (Full)& \cellcolor[rgb]{ .851,  .851,  .851}Task-Driven\\
& & &  &  & EA (VLM) & Way Points & EA (VLM) & URDF & Manipulation & Task Reward \\

\bottomrule
\end{tabular}%
\begin{tablenotes}    
        \footnotesize               
        \item[] \textbf{Note}: The \textcolor{magenta}{[link]} in each row directs to the corresponding code or project website. The shaded section indicates the category of each item, while the white section below lists the specific methods and representations used. For Latent-based methods, we further specify their generative models. \emph{\textbf{CL}: Closed-Loop, \textbf{OL}: Open-Loop, \textbf{F}: Fixed, \textbf{UF}: Unfixed, \textbf{Loco.}: Locomotion, \textbf{Grad. Opt.}: Gradient-Based Optimization, \textbf{Manip.}: Manipulation, \textbf{MCTS}: Monte‑Carlo Tree Search, \textbf{GHS}: Graph Heuristic Search, \textbf{MPPI}: Model Predictive Path Integral, \textbf{DE}: Differential Evolution, \textbf{GP}: Genetic Programming}
      \end{tablenotes}            
\end{threeparttable}       
} 
\label{tab:gen}
\vspace{-2mm}
\end{table*}    

\subsubsection{Latent-Based Generation}
Latent-based methods leverage neural generative models to learn smooth mappings from low-dimensional latent vectors to full 3D or graph-based morphologies. These learned latent spaces support efficient gradient-based, evolutionary, or probabilistic search over complex design spaces and are increasingly influential in ECD.

\emph{Variational Autoencoders} (VAEs) learn continuous latent representations of morphology and, in some cases, control parameters, facilitating sample-efficient optimization. For example, LABO~\cite{pan2021emergent} embeds high-dimensional hand morphologies and controller parameters into a VAE latent space and performs BO within that space, thereby reducing dimensionality and sample complexity for grasping tasks. GLSO~\cite{hu2023glso} extends this concept to graph-structured robots by training a graph-VAE on grammar-generated topologies and optimizing designs using BO. Liu et al.~\cite{liu2025terrain} proposed a Grammar-VAE and integrated differential evolution with a neural decoder, allowing the generative model to be reused across tasks. MorphVAE~\cite{song2024morphvae} further conditions latent sampling on task embeddings and introduces a \emph{continuous natural selection} strategy to improve diversity and performance.

\emph{Generative Adversarial Networks} (GANs) train a generator–discriminator pair to produce realistic morphologies by matching target data distributions. RoboGAN~\cite{hu2022modular} learns a one-to-many mapping from task descriptions (e.g., terrain type) to distributions over modular legged robots. Candidate bodies are sampled from the GAN manifold, while an evolutionary loop guides the search toward high-performing configurations, thereby enabling interpolation and diversity within the learned latent space.

\emph{Diffusion Models} have recently emerged as powerful generative priors for embodied co-design. DiffuseBot~\cite{wang2023diffusebot} integrates differentiable physics into the denoising process, using simulation feedback at each step to bias generation toward high-performing designs. Another recent work is Dynamics-Guided Diffusion Models (DGDM)~\cite{xu2024dynamics}, which replaces task-specific training data with physics-based guidance, enabling generalization to new manipulation tasks and novel geometries.

\emph{GFlowNets} offer an alternative generative formulation in which morphology–controller pairs are sampled with a probability proportional to their simulated performance. Recent works by \cite{nagiredla2024robonet}, \cite{nagiredlate}, and \cite{nagiredlacogent} have demonstrated that GFlowNets can generate diverse, high-quality designs while learning a structured policy over complete embodied agents.

Generative models can capture the inter-correlations among design variables. Once trained, a latent model can be reused across various tasks or paired with different controllers. However, training generative models typically requires large datasets and substantial simulation efforts. Moreover, latent spaces may decode to invalid or unrealistic morphologies without appropriate constraints. Achieving the right balance between fidelity and diversity remains a challenge.

\subsubsection{Large-Model-Based Generation}
Large pretrained models—including \emph{Large Language Models} (LLMs)~\cite{naveed2025comprehensive} and \emph{Vision–Language Models} (VLMs)~\cite{zhang2024vision}—have recently been introduced into ECD, by leveraging broad knowledge and powerful generative capabilities to propose, mutate, or refine morphologies and control strategies.

A growing body of work utilizes LLMs as intelligent mutation operators or search heuristics. A pioneering work is OpenELM~\cite{lehman2023evolution}, which demonstrates that a LLM trained on code can act as a high-quality mutation operator in genetic programming. In the \emph{Sodarace} experiments, the LLM consistently mutated code to generate robot walkers, and the resulting dataset was then used to train a conditional generative model that maps terrain descriptions to robot code. Building on this paradigm, LASeR~\cite{song2025laser} integrates a LLM into the optimization loop via a reflection mechanism that summarizes past search histories, prompting the LLM to generate improved design variants. RoboMoRe~\cite{fang2025robomore} further investigates LLM-driven co-design by instructing the LLM to propose candidate morphology–reward pairs jointly during a coarse search, after which promising candidates undergo alternating refinement. RoboMorph~\cite{qiu2024robomorph} represents each robot with a structured grammar and relies on LLM reasoning to iteratively improve morphologies through feedback.

LLMs have also been utilized for human–AI collaborative design. Stella et al.~\cite{stella2023can} have shown that a LLM can provide high-level conceptual guidance and partial specifications for a tomato-harvesting gripper, illustrating its potential in early-stage design. Beyond text-based reasoning, vision models are beginning to emerge in ECD. VLM-Engineer~\cite{gao2025vlmgineer} combines VLMs with evolutionary search to jointly design robotic tools and associated control policies. This line of work suggests future workflows in which a VLM first generates sketches or 2D concept designs from a task description, which are then translated into complete 3D morphologies.

Large models offer several advantages for ECD: they encapsulate extensive engineering and commonsense knowledge, enable creative and unconventional designs, support semantic prompting (e.g., ``design a robot capable of climbing a ladder’’), and diversify the search space. Their in-context learning capabilities make them effective interfaces between high-level human intent and concrete design proposals~\cite{makatura2023can,ma2024llm}. However, the outputs of LLMs and VLMs are not guaranteed to be feasible, optimal, or physically consistent, highlighting the need for stronger grounding and validation mechanisms.

\begin{table*}[htbp]
\renewcommand{\arraystretch}{0.9} 
\centering
\caption{Summary of Open-Ended Co-Design approaches.}
\vspace{-2mm}
\resizebox{\textwidth}{!}{
\begin{threeparttable} 
\begin{tabular}{c|l|c|c|c|ll|ll|ll}
\toprule
\multirow{2}[4]{*}{} & \multicolumn{1}{c|}{\multirow{2}[4]{*}{\textbf{Method}}} & \multicolumn{1}{c|}{\multirow{2}[4]{*}{\textbf{Year}}} & \multirow{2}[4]{*}{\textbf{\makecell{Agent\\Type}}} & \multirow{2}[4]{*}{\textbf{\makecell{Application\\(Sim/Real)}}} & \multicolumn{2}{c|}{\textbf{Controlling Brain}} & \multicolumn{2}{c|}{\textbf{Body Morphology}} & \multicolumn{2}{c}{\textbf{Task Environment}}\\
\cmidrule{6-11}      &       &       &       &       & \textbf{Design Space} & \textbf{Representation} & \textbf{Design Space} & \textbf{Representation}&\textbf{Design Space} & \textbf{Representation} \\
\midrule 
\multirow{4}[2]{*}{\rotatebox[origin=c]{90}{\textbf{Co-Evo}}} 
& \multirow{2}{*}{Stensby et al.~\cite{stensby2021co}} & \multirow{2}{*}{\makecell{2021\\~\href{https://github.com/EmmaStensby/poet-morphology}{[link]}}} & \multirow{2}{*}{\makecell{Legged\\Robot}} & \multirow{2}{*}{\makecell{Simulation\\\{Gym\}}} & \cellcolor[rgb]{ .851,  .851,  .851}Learnable (CL) & \cellcolor[rgb]{ .851,  .851,  .851}Individual & \cellcolor[rgb]{ .851,  .851,  .851}Continuous (F) & \cellcolor[rgb]{ .851,  .851,  .851}Direct & \cellcolor[rgb]{ .851,  .851,  .851}Modifiable (Full)& \cellcolor[rgb]{ .851,  .851,  .851}Hybrid \\
& & &  &  & EA (GA) & DNN & EA (GA) & Vector & Locomotion& Task+Diversity\\

\cmidrule{2-11}
& \multirow{2}{*}{MECE~\cite{ao2023curriculum}} & \multirow{2}{*}{\makecell{2023\\~\href{}{[---]}}} & \multirow{2}{*}{\makecell{Rigid\\Robot}} & \multirow{2}{*}{\makecell{Simulation\\\{MuJoCo\}}} & \cellcolor[rgb]{ .851,  .851,  .851}Learnable (CL) & \cellcolor[rgb]{ .851,  .851,  .851}Universal & \cellcolor[rgb]{ .851,  .851,  .851}Hybrid (UF) & \cellcolor[rgb]{ .851,  .851,  .851}Indirect & \cellcolor[rgb]{ .851,  .851,  .851}Modifiable (Full)& \cellcolor[rgb]{ .851,  .851,  .851}Task-Driven\\
& & &  &  & RL (PPO) & GNN & RL (PPO) & Graph (GNN) & Locomotion& Task Reward\\

\midrule
\multirow{3}[2]{*}{\rotatebox[origin=c]{90}{\textbf{Developmental Methods}}} 

& \multirow{2}{*}{Rieffel et al.~\cite{rieffel2014growing}} & \multirow{2}{*}{\makecell{2014\\~\href{}{[---]}}} & \multirow{2}{*}{\makecell{Soft\\Robot}} & \multirow{2}{*}{\makecell{Simulation \\\{PhysX\}}} & \cellcolor[rgb]{ .851,  .851,  .851}Learnable (OL) & \cellcolor[rgb]{ .851,  .851,  .851}Individual & \cellcolor[rgb]{ .851,  .851,  .851}Hybrid (UF) & \cellcolor[rgb]{ .851,  .851,  .851}Indirect & \cellcolor[rgb]{ .851,  .851,  .851}Predefined (Full)& \cellcolor[rgb]{ .851,  .851,  .851}Task-Driven\\
& & &  &  & EA (GA) & CPG & EA (GA) & L-System & Locomotion& Task Reward\\


\cmidrule{2-11}
& \multirow{2}{*}{Joachimczak et al.~\cite{joachimczak2016artificial}} & \multirow{2}{*}{\makecell{2016\\~\href{}{[---]}}} & \multirow{2}{*}{\makecell{Virtual\\Creature}} & \multirow{2}{*}{\makecell{Simulation \\\{Customized\}}} & \cellcolor[rgb]{ .851,  .851,  .851}Learnable (OL) & \cellcolor[rgb]{ .851,  .851,  .851}Universal & \cellcolor[rgb]{ .851,  .851,  .851}Hybrid (UF) & \cellcolor[rgb]{ .851,  .851,  .851}Indirect & \cellcolor[rgb]{ .851,  .851,  .851}Predefined (Full)& \cellcolor[rgb]{ .851,  .851,  .851}Task-Driven\\
& & &  &  & EA (NEAT) & ANN\&CPG & EA (NEAT) & ANN & Locomotion& Task Reward\\

\cmidrule{2-11}
& \multirow{2}{*}{Vujovic et al.~\cite{vujovic2017evolutionary}} & \multirow{2}{*}{\makecell{2017\\~\href{}{[---]}}} & \multirow{2}{*}{\makecell{Legged\\Robot}} & \multirow{2}{*}{\makecell{Real-World \\\{---\}}} & \cellcolor[rgb]{ .851,  .851,  .851}Learnable (OL) & \cellcolor[rgb]{ .851,  .851,  .851}Individual & \cellcolor[rgb]{ .851,  .851,  .851}Hybrid (F) & \cellcolor[rgb]{ .851,  .851,  .851}Direct & \cellcolor[rgb]{ .851,  .851,  .851}Predefined (Full)& \cellcolor[rgb]{ .851,  .851,  .851}Task-Driven\\
& & &  &  & EA & PID & EA & Vector & Locomotion& Task Reward\\

\cmidrule{2-11}
& \multirow{2}{*}{Naya et al.~\cite{naya2020experiment}} & \multirow{2}{*}{\makecell{2020\\~\href{}{[---]}}} & \multirow{2}{*}{\makecell{Quadrupedal \\Robot}} & \multirow{2}{*}{\makecell{Simulation \\\{V-REP\}}} & \cellcolor[rgb]{ .851,  .851,  .851}Learnable (OL) & \cellcolor[rgb]{ .851,  .851,  .851}Individual & \cellcolor[rgb]{ .851,  .851,  .851}Continuous (F) & \cellcolor[rgb]{ .851,  .851,  .851}Direct & \cellcolor[rgb]{ .851,  .851,  .851}Predefined (Full)& \cellcolor[rgb]{ .851,  .851,  .851}Task-Driven\\
& & &  &  & EA (NEAT) & ANN & Development & Vector & Locomotion& Task Reward\\

\cmidrule{2-11}
& \multirow{2}{*}{Benureau et al.~\cite{benureau2022morphological}} & \multirow{2}{*}{\makecell{2022\\~\href{}{[---]}}} & \multirow{2}{*}{\makecell{Virtual\\Creature}} & \multirow{2}{*}{\makecell{Simulation \\\{Customized\}}} & \cellcolor[rgb]{ .851,  .851,  .851}Learnable (OL) & \cellcolor[rgb]{ .851,  .851,  .851}Individual & \cellcolor[rgb]{ .851,  .851,  .851}Continuous (F) & \cellcolor[rgb]{ .851,  .851,  .851}Direct & \cellcolor[rgb]{ .851,  .851,  .851}Predefined (Full)& \cellcolor[rgb]{ .851,  .851,  .851}Task-Driven\\
& & &  &  & EA (ES) & CPG & Development & Vector &Locomotion & Task Reward\\

\cmidrule{2-11}
& \multirow{2}{*}{NCRS~\cite{pontes2022single}} & \multirow{2}{*}{\makecell{2022\\~\href{https://github.com/sidneyp/neural-cellular-robot-substrate}{[link]}}} & \multirow{2}{*}{\makecell{Modular\\Robot}} & \multirow{2}{*}{\makecell{Simulation \\\{Box2d\}}} & \cellcolor[rgb]{ .851,  .851,  .851}Learnable (CL) & \cellcolor[rgb]{ .851,  .851,  .851}Individual & \cellcolor[rgb]{ .851,  .851,  .851}Discrete (UF) & \cellcolor[rgb]{ .851,  .851,  .851}Indirect & \cellcolor[rgb]{ .851,  .851,  .851}Predefined (Partial)& \cellcolor[rgb]{ .851,  .851,  .851}Task-Driven\\
& & &  &  & EA (CMA-ES) & NCA & EA (CMA-ES) & NCA & Navigation& Task Reward\\

\cmidrule{2-11}
& \multirow{2}{*}{Higashinaka et al.~\cite{higashinaka2024evolution}} & \multirow{2}{*}{\makecell{2024\\~\href{}{[---]}}} & \multirow{2}{*}{\makecell{Modular\\Soft Robot}} & \multirow{2}{*}{\makecell{Simulation \\\{EvoGym\}}} & \cellcolor[rgb]{ .851,  .851,  .851}Predefined (OL) & \cellcolor[rgb]{ .851,  .851,  .851}Individual & \cellcolor[rgb]{ .851,  .851,  .851}Discrete (UF) & \cellcolor[rgb]{ .851,  .851,  .851}Direct & \cellcolor[rgb]{ .851,  .851,  .851}Predefined (Full)& \cellcolor[rgb]{ .851,  .851,  .851}Task-Driven\\
& & &  &  & None & CPG & EA (GA) & Matrix & Locomotion &Task Reward \\

\cmidrule{2-11}
& \multirow{2}{*}{Jaafar et al.~\cite{jaafar2025eco}} & \multirow{2}{*}{\makecell{2025\\~\href{}{[---]}}} & \multirow{2}{*}{\makecell{Virtual\\Creature}} & \multirow{2}{*}{\makecell{Simulation \\\{Bullet\}}} & \cellcolor[rgb]{ .851,  .851,  .851}Learnable (CL) & \cellcolor[rgb]{ .851,  .851,  .851}Individual & \cellcolor[rgb]{ .851,  .851,  .851}Hybrid (UF) & \cellcolor[rgb]{ .851,  .851,  .851}Indirect & \cellcolor[rgb]{ .851,  .851,  .851}Modifiable (Partial)& \cellcolor[rgb]{ .851,  .851,  .851}Task-Driven\\
& & &  &  & HyperNEAT & CPPN\&DNN & HyperNEAT & CPPN & Locomotion &Task Reward \\

\bottomrule
\end{tabular}%
\begin{tablenotes} 
\footnotesize               
\item[] \textbf{Note}: The \textcolor{magenta}{[link]} in each row directs to the corresponding code or project website. The shaded section indicates the category of each item, while the white section below lists the specific methods and representations used. \emph{\textbf{CL}: Closed-Loop, \textbf{OL}: Open-Loop, \textbf{F}: Fixed, \textbf{UF}: Unfixed, \textbf{Loco.}: Locomotion, \textbf{Manip.}: Manipulation, \textbf{PID}: Proportional Integral Derivative, \textbf{NCA}: Neural Cellular Automata}
\end{tablenotes} 
\end{threeparttable}       
} 
\label{tab:open}
\vspace{-2mm}
\end{table*}


\subsection{Open-Ended Co-Design}\label{open}
Open-ended co-design aims to achieve continual improvement by allowing an agent’s morphology, controller, and environment to co-evolve. By removing fixed terminal objectives, these systems foster sustained innovation, as agents and environments mutually generate new challenges and adaptations, as shown in Fig.~\ref{fig_3} (d). We categorize open-ended co-design into two primary subtopics: 1) \emph{\textbf{Brain–Body–Environment Co-Evolution}} and 2) \emph{\textbf{Developmental Evolution}}.

\subsubsection{Brain-Body-Environment Co-Evolution}
This approach treats the task environment as a co-optimized component alongside morphology and control. Foundational work in this direction is the \emph{Paired Open-Ended Trailblazer} (POET) framework~\cite{wang2019poet,wang2020enhanced}. POET maintains a population of environment–controller pairs, simultaneously mutating environments and evolving agents. Policies that succeed in one environment may be transferred to other contexts. Stensby et al.~\cite{stensby2021co} extended POET to jointly evolve morphologies and controllers in tandem with their environments. Their results indicate that learned environmental progressions yield greater morphological diversity than handcrafted curricula and reduce the risk of premature convergence.

Recent work has introduced RL-driven co-evolution of the brain, body, and environment. Ao et al. proposed MECE~\cite{ao2023curriculum}, which formulates a curriculum where both morphology and the environment evolve according to learned RL policies. This approach produces progressively challenging and structured tasks that promote sustained adaptation. Another emerging trend enhances environment evolution through the use of generative models. For instance, LLM-POET~\cite{aki2024llm} replaces manual environment mutation with a LLM that generates increasingly complex environments for a population of soft-robot agents. As agents improve, the LLM proposes novel tasks that escalate in difficulty, creating an open-ended loop of continual challenge and innovation.

\subsubsection{Developmental Evolution}
Developmental co-evolution draws inspiration from evolutionary developmental biology (Evo-Devo) and developmental robotics, where genotypes encode growth processes that shape both morphology and control over an agent’s lifetime~\cite{kriegman2018morphological,van2021comparing,arita2016alife,bongard2011morphological,bongard2010utility}. Early research has investigated developmental encodings, which include L-systems~\cite{rieffel2014growing}, grammar rules, and neural networks~\cite{pontes2022single,joachimczak2016artificial}. Bongard~\cite{bongard2003evolving} demonstrated that \emph{artificial ontogeny} can generate hierarchical, repeated structures and controllers that emerge as bodies grow, yielding more structured solutions. 

The other research focus is on morphological development. Here, developmental rules adjust parameters, such as leg angles or actuator amplitudes, during learning~\cite{naya2020experiment} or progressively ``unfreeze” degrees of freedom to facilitate skill acquisition~\cite{lungarella2002adaptivity,berthouze2004motor,savastano2012incremental}. Later work introduced mechanisms for growing new body parts during an agent’s lifetime~\cite{benureau2022morphological,jaafar2025eco,vujovic2017evolutionary}, providing additional adaptive flexibility. For comprehensive overviews of developmental evolution, we refer readers to~\cite{naya2021morphological,wang2021survey}. Overall, open-ended co-design is not yet a unified methodology but rather a collection of emerging ideas that leverage ontogeny, plasticity, and staged learning, and it awaits further exploration.

\section{Testbeds: Applying Embodied Co-Design}\label{benapp}
Embodied co-design has evolved rapidly due to the availability of high-fidelity simulation platforms and increasingly advanced real-world fabrication techniques. These resources provide standardized environments, reproducible tasks, and practical feedback loops that facilitate the automated design and validation of embodied agents. This section surveys representative benchmarks, datasets, and applications of ECD across both simulated and real-world domains.

\subsection{Simulated Platforms}
\subsubsection{Rigid-Body Simulators} As summarized in Table~\ref{tab:platforms}, widely used robot simulators, including MuJoCo~\cite{todorov2012mujoco}, Bullet~\cite{panerati2021learning}, Unity~\cite{juliani2018unity,Veenstra2023}, Gazebo~\cite{koenig2004design}, Webots~\cite{michel2004cyberbotics}, SAPIEN~\cite{xiang2020sapien}, DART~\cite{lee2018dart}, Rostok~\cite{borisov2022reconfigurable}, and V-Rep~\cite{rohmer2013v}, provide native support for standard robot description formats such as XML, MJCF, URDF, and SDF. This compatibility facilitates rapid modifications to morphologies and establishes these simulators as foundational tools for numerous works in ECD. The convergence of robotics and AI has further increased the demand for scalable, GPU-accelerated simulations. Platforms such as NVIDIA Isaac Gym and its extensions~\cite{makoviychuk2021isaac, kulkarni2023aerial,gu2024humanoid} enable massively parallel environments that support simultaneous training of large sets of morphological variants with controllers~\cite{cheng2024structural,wang2024evolution,chen2023evolving}. 

While much of the existing ECD literature relies on non-differentiable simulators that only provide forward dynamics, a growing number of differentiable platforms supporting inverse dynamics have emerged. These tools—Dojo~\cite{howell2022dojo}, Brax~\cite{freeman2021brax}, Neuralsim~\cite{heiden2021neuralsim}, Nimble~\cite{werling2021fast} (the differentiable version of DART), DiffRedMax~\cite{xu2021accelerated}, and SDRS~\cite{ye2024sdrs}—facilitate physics-based co-design. Building on these efforts, frameworks such as Genesis~\cite{Genesis} and MuJoCo Playground~\cite{mujoco_playground_2025} aim to unify differentiable physics, GPU acceleration, and integrated rendering within a single API.

\subsubsection{Soft-Body Simulators} For soft-bodied agents, early platforms such as SOFA~\cite{coevoet2017software,deimel2017automated}, VoxCAD~\cite{cheney2013unshackling,cheney2014electro,cheney2018scalable}, and VoxCraft~\cite{kriegman2021fractals,li2024reinforcement} remain influential due to their detailed physical models of deformable materials. Lightweight two-dimensional frameworks, such as 2D-VSR-Sim~\cite{ferigo2025totipotent}, further enable rapid algorithmic prototyping. Recently, EvolutionGym~\cite{bhatia2021evolution} introduces a comprehensive suite of tasks along with a diverse material palette for voxel-based soft robots, which has since been widely adopted in ECD research~\cite{katayama2023ga,dongleveraging2024,harada2024lamarckian,zhao2024morphological,xie2025accelerating,zhao2025multi,liu2023rapidly,saito2024effective,zhao2025cross,davis2023subtract,tanaka2022co,mertan2024no,mertan2025controller,nadizar2025enhancing,song2025laser,song2024morphvae,liu2025morphology}. Complementary to voxel-based systems, several platforms integrate differentiable soft-body physics, including DiffTaichi~\cite{hu2019difftaichi}, ChainQueen~\cite{hu2019chainqueen}, and DiffAqua~\cite{ma2021diffaqua}. SoftZoo~\cite{wang2023softzoo} advances this direction by providing a benchmark across various terrains (e.g., flat ground, snow, and water) and task objectives (e.g., speed, turning, and path following). In parallel, Sorotoki~\cite{caasenbrood2024sorotoki} provides an integrated MATLAB toolkit.

\subsubsection{Embodied Tasks Simulators} Notably, recent specialized simulators for embodied intelligence, such as iGibson~\cite{li2022igibson}, AI2-THOR~\cite{kolve2017ai2}, Habitat-Sim~\cite{szot2021habitat}, and LIBERO~\cite{liu2023libero}, provide realistic and interactive 3D environments that integrate rich visual, physical, and semantic information. These platforms support the training and evaluation of increasingly complex embodied agents, including those with advanced control paradigms (e.g., Vision-Language-Action models~\cite{zitkovich2023rt,kim2024openvla,bjorck2025gr00t}) and various morphologies (e.g., humanoids). Furthermore, they provide challenging benchmarks for long-horizon tasks involving navigation, manipulation, and multi-step planning, which makes them well-suited foundations for advancing ECD research in the future.

\begin{table}[t]
  \centering
  \caption{A Summary of the Past Decade’s Simulators and Physics Engines Supporting Embodied Co-Design.}
  \vspace{-2mm}
  \resizebox{0.46\textwidth}{!}{
  \begin{tabular}{@{}l|l|c|c|c|c@{}}
    \toprule
     & \textbf{Platform} & \textbf{Year} & \textbf{Agent} & \textbf{Differentiable} & \textbf{Link} \\
    \midrule

      & SOFA & 2015 & Soft  & No & \href{https://www.sofa-framework.org/}{[link]} \\
      & VoxCAD & 2015 & Soft  & No & \href{https://www.creativemachineslab.com/voxcad.html}{[link]} \\
      & ChainQueen & 2018 & Soft  & Yes & \href{https://github.com/yuanming-hu/ChainQueen}{[link]} \\
      & DART & 2018 & Rigid  & No & \href{https://github.com/dartsim/dart}{[link]} \\
      & Webots & 2018 & Mixed  & No & \href{https://github.com/cyberbotics/webots}{[link]} \\
      & DiffTaichi & 2019 & Mixed  & Yes & \href{https://github.com/taichi-dev/difftaichi}{[link]} \\
      & Revolve & 2019 & Soft  & No & \href{https://github.com/ci-group/revolve}{[link]} \\
      & Habitat-Sim & 2019 & Mixed  & No & \href{https://github.com/facebookresearch/habitat-sim}{[link]} \\
      & SAPIEN & 2020 & Rigid & No & \href{https://github.com/haosulab/SAPIEN}{[link]} \\
      & 2D-VSR-Sim & 2020 & Soft & No & \href{https://github.com/ElsevierSoftwareX/SOFTX_2020_14}{[link]} \\
      & TDS & 2020 & Rigid & Yes & \href{https://github.com/erwincoumans/tiny-differentiable-simulator}{[link]} \\
      & Brax & 2021 & Rigid & Yes & \href{https://github.com/google/brax}{[link]} \\
      & MuJoCo & 2021 & Mixed & No & \href{https://mujoco.org/}{[link]} \\
      & PyBullet & 2021 & Mixed & No & \href{https://pybullet.org/wordpress/}{[link]} \\
      & Isaac Gym & 2021 & Mixed & No & \href{https://developer.nvidia.com/isaac-gym}{[link]} \\
      & Evolution Gym & 2021 & Soft & No & \href{https://evolutiongym.github.io/}{[link]} \\
      & Nimble Physics & 2021 & Rigid & Yes & \href{https://nimblephysics.org/}{[link]} \\
      & DiffHand & 2021 & Rigid & Yes & \href{https://github.com/eanswer/DiffHand}{[link]} \\
      & DiffAqua & 2021 & Soft & Yes & \href{https://github.com/mit-gfx/DiffAqua}{[link]} \\
      & AI2-THOR & 2021 & Mixed & No & \href{https://github.com/allenai/ai2thor}{[link]} \\
      & iGibson & 2021 & Mixed & No & \href{https://github.com/StanfordVL/iGibson}{[link]} \\
      & Modular EvoGym & 2022 & Soft & No & \href{https://github.com/Yuxing-Wang-THU/ModularEvoGym}{[link]} \\
      & 2D-MR-Sim & 2022 & Soft & No & \href{https://github.com/ericmedvet/2dmrsim}{[link]} \\
      & 2D-Robot-Evolution & 2022 & Soft & No & \href{https://github.com/ericmedvet/2d-robot-evolution}{[link]} \\
      & Rostok & 2022 & Mixed & No & \href{https://github.com/aimclub/rostok}{[link]} \\
      & Dojo & 2022 & Rigid & Yes & \href{https://github.com/dojo-sim/Dojo.jl}{[link]} \\
      & EMR & 2023 & Modular & No & \href{https://github.com/FrankVeenstra/EvolvingModularRobots_Unity}{[link]} \\
      & SoftZoo & 2023 & Soft & Yes & \href{https://github.com/zswang666/softzoo}{[link]} \\
      & LIBERO & 2023 & Rigid & No & \href{https://github.com/Lifelong-Robot-Learning/LIBERO}{[link]} \\
      & RoboGen & 2024 & Mixed & No & \href{https://robogen-ai.github.io/}{[link]} \\
      & Revolve2 & 2024 & Mixed & No & \href{https://ci-group.github.io/revolve2/}{[link]} \\
      & Genesis & 2024 & Mixed & Yes & \href{https://github.com/Genesis-Embodied-AI/Genesis}{[link]} \\
      & Sorotoki & 2024 & Soft& No & \href{https://bjcaasenbrood.github.io/SorotokiCode/}{[link]} \\
      & SDRS & 2024 & Soft & Yes & \href{https://bitbucket.org/runningblade/libdifferentiable/src/ConvexHullPBAD/}{[link]} \\
    \bottomrule
  \end{tabular}
  }
  \label{tab:platforms}
  \vspace{-2mm}
\end{table}

\subsection{Datasets} In addition to the aforementioned simulation platforms, several ECD datasets have been developed to support the systematic analysis of agent design. RoboCrafter-QA~\cite{chen2025large} provides comparative data to evaluate whether LLMs can effectively link high-level task descriptions to low-level morphological and material choices in soft robotics. RoboDesign1M~\cite{le2025robodesign1m} offers a large-scale dataset of one million robotic design samples across multiple domains, enabling tasks such as design–image generation, design-centric visual question answering, and design–image retrieval.    

\begin{figure*}[t]
\centering
\includegraphics[width=\textwidth]{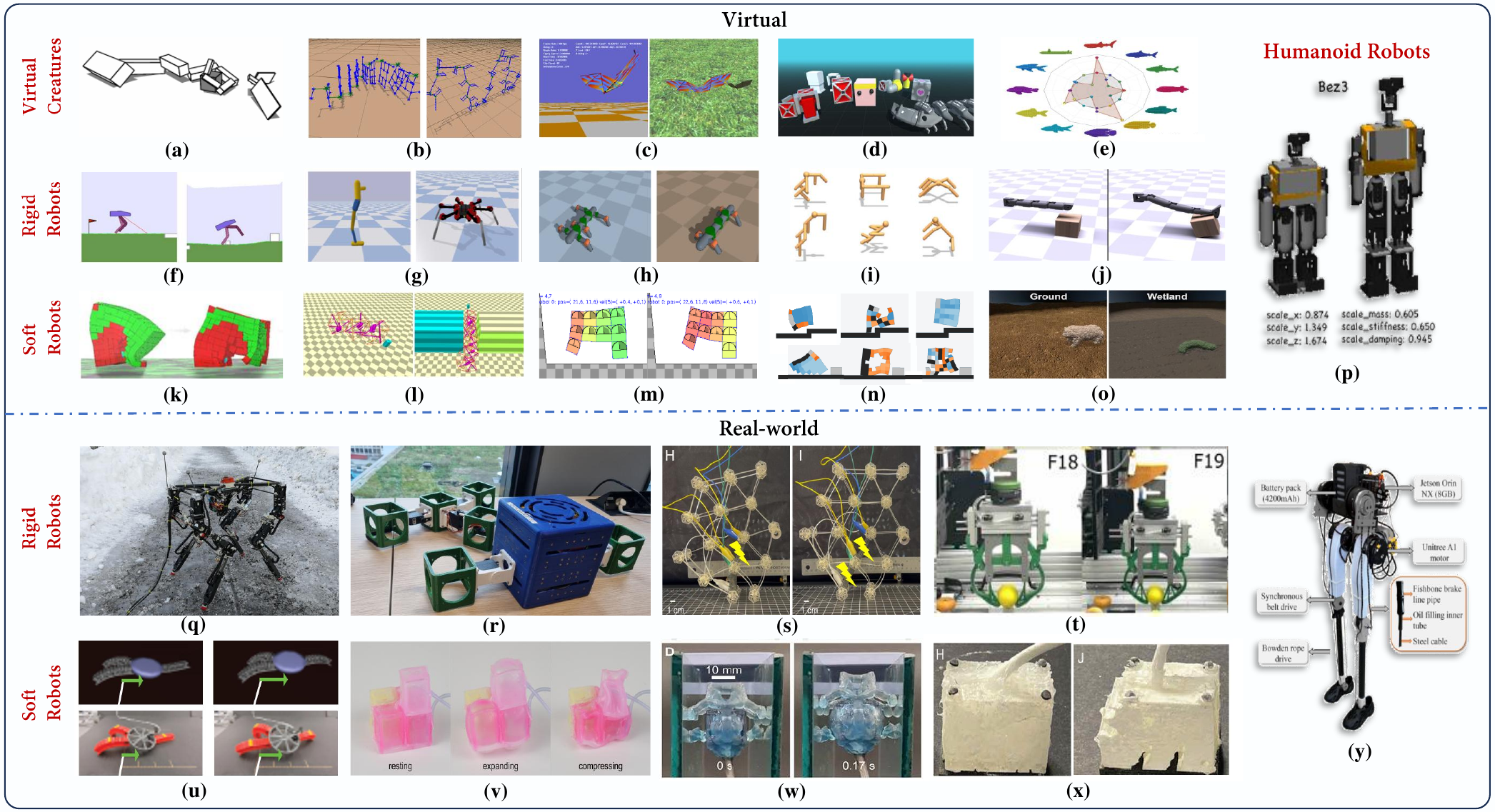}
\caption{Co-designed embodied agents in both virtual and real environments. Example images numbered from (a) to (p) are adapted from \cite{sims1994competition,hornby2001evolving,shim2003generating,Veenstra2023,ma2021diffaqua,ha2019reinforcement,luck2020data,zhao2020robogrammar,gupta2021embodied,xu2021accelerated,cheney2013unshackling,zardini2021seeking,medvet2021biodiversity,wang2023curriculum,wang2023softzoo,yue2025toward}, respectively. Example images numbered from (q) to (y) are adapted from \cite{nygaard2021environmental,revolve2,Strgar-RSS-24,xie2024map,schaff2022soft,liu_voxcraft_2020,kobayashi2024computational,matthews2023efficient,cheng2024structural}, respectively.}
\label{fig_4}
\end{figure*}

\subsection{Real-World Applications}\label{realworld}
While simulation facilitates rapid exploration, physical validation remains the ultimate test of ECD. Recent advances in fabrication, 3D printing, and embedded sensing have made it feasible to realize co-designed morphologies. In this section, we summarize recent real-world applications (Fig.~\ref{fig_4}).

\subsubsection{Rigid Robots} These efforts primarily focus on developing agents with enhanced stability, agility, and energy efficiency. Many systems adopt bi-level or single-level co-design frameworks to jointly optimize morphology and control. Recent real-world demonstrations span bipedal~\cite{cheng2024structural}, quadrupedal~\cite{bjelonic2023learning,bravo2024engineering,nygaard2021environmental,bravo2024engineering,ringel2025text2robot,ha2018computational2,ha2017joint}, micro-scale~\cite{spielberg2017functional}, and wheeled robots~\cite{park2021computational}, resulting in highly adaptable agents that demonstrate robust performance.

\subsubsection{Soft Robots} Researchers have also investigated the embodied co-design of soft robots made from highly deformable materials~\cite{caasenbrood2024sorotoki}. A pioneering example is the shape-changing crawling robot introduced in~\cite{schaff2022soft}, which demonstrates how deformable morphologies can be co-optimized with control for adaptive locomotion. In parallel, the VoxCraft platform~\cite{kriegman2021fractals} enables the rapid physical realization of voxel-based soft robots through modular assembly and disassembly, thereby supporting large-scale embodied evolution in hardware-rich environments. Among these applications, we also observe emerging trends in leveraging differentiable simulation alongside advanced actuation methods to narrow the co-design gap between simulation and real-world deployment~\cite{matthews2023efficient,bacher2021design}.   
 
\subsubsection{Modular Robot} Open frameworks, such as Revolve~\cite{revolve2,lan2021learning} and RoboGen~\cite{auerbach2014robogen,jelisavcic2017real}, provide C++ and Python libraries that facilitate the creation, simulation, and management of robots within the Gazebo platform. These frameworks also enable virtual designs to be 3D-printed and subsequently deployed in the real world. More recently, the Granulobot system~\cite{saintyves2024self} demonstrates how gear-like modular units can self-assemble into aggregates with tunable rigidity, offering new insights into ECD. Similarly, EMARGE~\cite{moreno2021emerge} introduces a platform for full-cycle morphological evolution in reconfigurable hardware. We refer readers to \cite{liang2025decoding} for a comprehensive review along this line.

\subsubsection{Aerial and Aquatic Vehicles} For aerial agents, ECD targets aerodynamic efficiency, control efficiency, and payload trade-offs, which are typically validated through flight testing or wind-tunnel experiments~\cite{bergonti2024co,xu2019learning}. Aquatic agents—both underwater and surface—require joint optimization of geometry, propulsion, and control to reduce drag and improve maneuverability and sensing in fluid environments, with evaluations conducted in pools or through field deployments~\cite{katzschmann2018exploration, sariman2025ur}.

\subsubsection{Surgical Robots} ECD is also gaining increasing attention in surgical robotics and wearable device design, where co-optimization must simultaneously satisfy strict safety, sterility, and human-compatibility requirements~\cite{berger2022design,alcaraz2024using,alcaraz2025designing}.

\subsubsection{Humanoids} Embodied co-design for humanoids focuses on anthropomorphic morphology, joint ranges, balance, and perception–action integration, all under strict safety and ergonomic constraints~\cite{ghansah2023humanoid,vanteddu2024cad,liu2025embracing}. 

\subsubsection{Manipulators and End-Effectors} Because full-body real-world ECD remains highly challenging, much practical progress has been made in co-designing manipulators and specialized end-effectors~\cite{chen2021co}. Approaches include generative methods based on diffusion models~\cite{wang2023diffusebot,xu2024dynamics}, Grammar-based search~\cite{zharkov2024synergizing}, bi-level co-design with surrogate models~\cite{bai2025learning,islam2024task,dong2024cagecoopt}, and single-level co-design leveraging learned dynamics models~\cite{yi2025co}. These methods yield manipulators and tools with improved efficiency, robustness, and generalization in real-world manipulation tasks.

\section{Challenges and Future Trends}\label{futu}
In this section, we identify some open challenges and future directions in ECD, based on both retrospectives of the methods discussed in this survey and outlooks to the emerging trends of embodied intelligence.

\subsection{Open Challenges}
\subsubsection{Efficiency and Scalability} The co-design space of embodied agents is prohibitively vast. As discussed in Section~\ref{bi-level}, a wide range of ECD methods typically require expensive inner-loop policy optimization processes to support the morphological search. This computational burden makes the development of efficient and accurate performance predictors essential for scaling ECD to complex, high-dimensional scenarios. In addition, although techniques such as curriculum learning~\cite{wang2023curriculum,ao2023curriculum}, representation learning~\cite{spielberg2019learning,song2024morphvae}, and transfer learning~\cite{zhao2024morphological,harada2024lamarckian,zhao2025cross,chen2018hardware} have been introduced to promote co-design knowledge reuse in ECD, scalability remains a critical challenge. Co-design approaches need to generalize across agents of varying sizes and structural complexities, rather than being limited to narrow task domains or specific morphological scales.

\subsubsection{Generalization} Most ECD techniques follow a ``one-robot–one-task” paradigm, optimizing morphology and control specifically for a single task. This contrasts with biological organisms—such as humans—whose morphologies enable a broad repertoire of behaviors without task-specific redesign. A key challenge is the explicit co-design of \emph{generalist} agents capable of performing multiple tasks, rather than narrowly tuned specialists~\cite{wang2023preco}. Advancing toward such generality raises complementary challenges, including the design of agents with intrinsic structural resilience, self-assessment, and self-repair capabilities~\cite{horibe2021regenerating,bilodeau2017self,islam2023review,xie2024soft}, and methods that support the \emph{online} co-adaptation of morphology and control in response to changing environmental conditions.

\subsubsection{Sim-to-Real} As introduced in Section~\ref{realworld}, despite significant advances in simulation, closing the sim-to-real gap remains a major challenge for embodied co-design~\cite{eiben2021real,kriegman2020scalable,van2021influence}. Morphologies that perform well under idealized physics often fail in reality due to unmodeled factors such as frictional variability, material nonlinearity, complex contact dynamics, heterogeneity, sensor noise, actuation delays, and manufacturing imperfections. A widely adopted strategy is to perform the morphological search on cheap surrogates~\cite{he2024morph,du2025efficient}, followed by validation and refinement in high-fidelity simulation or on hardware, incorporating human manufacturability checks before fabrication. Emerging techniques, including domain randomization~\cite{fadini2022simulation,schaff2023sim,lirobottool2023} and differentiable simulation with realistic material models~\cite{bacher2021design}, aim to further mitigate this gap. From the fabrication perspective, realizing evolved morphologies—especially multi-material, non-planar, embedded-sensor assemblies, and highly deformable structures—remains challenging. Promising avenues for narrowing the gap include digital twins for tighter sim–real integration and reconfigurable modular hardware that enables rapid physical validation and iterative co-design.

\subsubsection{Theories} Although numerous ECD approaches have emerged, few works have explored the theoretical foundations underlying this approach. A solid theoretical foundation can not only guide the design of more effective embodied agents but also clarify fundamental concepts, such as the \emph{brain-body trade-off}. Furthermore, more practical metrics grounded in morphological computation could offer a holistic assessment of agent performance and prevent the unprincipled replacement of components with incremental technologies in common ECD pipelines. Hence, a more unified theory is indispensable to take the development of ECD to the next stage.

\subsubsection{Benchmarks and Datasets} 
The ECD field still lacks widely adopted benchmarks, though platforms such as EvolutionGym~\cite{bhatia2021evolution} and SoftZoo~\cite{wang2023softzoo} represent important initial steps. Unlike traditional control optimization problems, benchmarking embodied co-design poses unique challenges, requiring diverse metrics to reflect the coupled nature of morphology and control. Expanding the infrastructure and establishing standardized ECD datasets could facilitate more rigorous and reproducible evaluations in ECD research.
 
\subsection{Future Trends}
Despite the challenges mentioned above in ECD, many promising directions remain worth exploring. From our perspective, future trends in ECD may include, but are not limited to, the following three main topics:

\subsubsection{Human-AI Collaborative ECD} Rather than replacing humans, generative AI tools—such as LLMs and VLMs—can act as co-designers by proposing morphologies, aligning semantic descriptions with agent behaviors~\cite{matthews2020crowd}, explaining trade-offs, drafting fabrication steps, and generating initial controllers based on high-level objectives~\cite{stella2023can,chen2025robots}. Meanwhile, human designers contribute domain knowledge, safety judgments, and ethical priors. Combining both approaches reduces design cycle time, increases the diversity of candidate designs, and makes ECD accessible to non-specialists. Recent works~\cite{song2025laser,ringel2025text2robot,fang2025robomore,gao2025vlmgineer,kulz2025design} have demonstrated preliminary pipelines that can transform natural language prompts into robot designs within minutes to hours. However, the full potential of these tools in reasoning and explainability to inform human designers has yet to be explored. Therefore, we believe that investigating such a human-ai pipeline is a promising direction.

\subsubsection{Multi-Agent Collaborative ECD} This direction envisions a framework in which multiple autonomous machines—such as software agents, specialized LLMs, or learned policies—collaboratively design an embodied robot. Each agent contributes distinct capabilities, enabling interaction modes such as coordination, collaboration, or even competition. For example, a ``morphology agent” proposes geometry, an ``actuation agent” selects motors or stiffness parameters, a ``control agent” designs controllers, and a ``fabrication agent” ensures manufacturability. Together, these agents coordinate to generate coherent robot designs. Recent \emph{agentic} frameworks have provided practical blueprints for such distributed co-design~\cite{shen2023hugginggpt,yang2024llm,qiu2024llm}. By decomposing the ECD problem into specialized roles, multi-agent systems reduce per-agent complexity, enable parallel exploration, and align naturally with modular hardware development pipelines.

\subsubsection{Open-Ended and Lifelong ECD} As discussed in Section~\ref{open}, open-ended co-design enables systems to continuously generate new tasks while concurrently optimizing morphologies and control policies to address them. This suggests that, when combined with lifelong-learning techniques~\cite{liu2021lifelong,mendez2023embodied,liu2023libero} that accumulate, reuse, and compose skills over long horizons, such systems can produce an ongoing stream of increasingly capable embodied machines. ECD, therefore, has the potential to evolve into a persistent creative ecology rather than remaining a bounded optimization problem.

\section{Conclusion}
The interplay among morphology, control, and task environment is fundamental not only for biological organisms but also for the development of intelligent embodied agents. In this survey, we review recent advances in embodied co-design (ECD) and organize them into a hierarchical taxonomy. This taxonomy provides a structured framework that enables researchers to contextualize current progress and identify opportunities for cross-disciplinary collaboration. We also review notable applications in both simulated and physical environments. Although the current results indicate that co-designed agents outperform fixed-body baselines, many questions remain unanswered. We highlight these challenges and discuss promising directions for future research. We hope this survey serves as both a foundational reference for newcomers and a comprehensive resource for experts seeking deeper insights into ECD.

\bibliographystyle{IEEEtran}
\bibliography{reference.bib}

\begin{IEEEbiography}[{\includegraphics[width=1in,height=1.25in,clip,keepaspectratio]{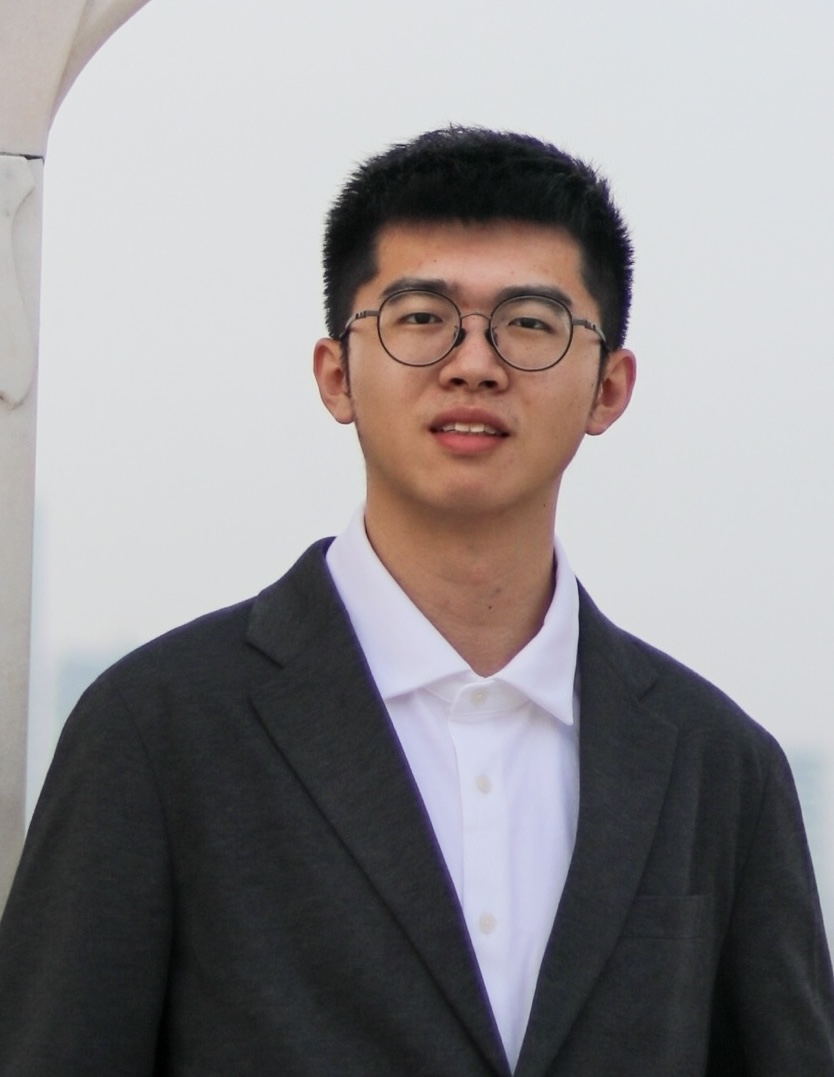}}]{Yuxing Wang}
(Member, IEEE) received the B.Eng. degree in communication engineering from Southwest Minzu University (SMU) in 2020, and the M.Eng. degree in electronic and information engineering from Tsinghua University (THU) in 2023.

He is currently pursuing a Ph.D. degree in control science and engineering at THU. Before that, he worked as a research assistant at THU and as a research intern at Tencent AI Lab. His research interests include evolutionary learning and embodied intelligence, with a particular focus on evolutionary reinforcement learning for control optimization, and embodied co-design for modular robotic systems.
\end{IEEEbiography}

\begin{IEEEbiography}[{\includegraphics[width=1in,height=1.25in,clip,keepaspectratio]{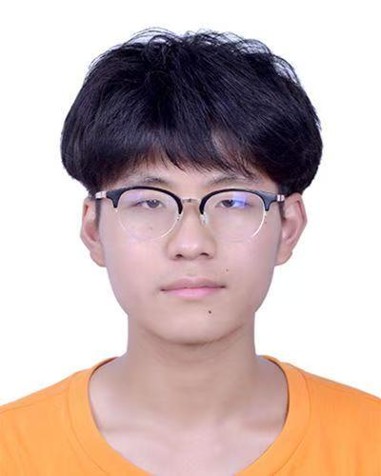}}]{Zhiyu Chen} received the B.Sc. degree in electronic engineering from Tsinghua University, Beijing, China, in 2025. He is currently pursuing a M.Eng. degree in artificial intelligence at Tsinghua University. His research interests are in reinforcement learning theory, embodied intelligence, and large language models.
\end{IEEEbiography}

\begin{IEEEbiography}[{\includegraphics[width=1in,height=1.25in,clip,keepaspectratio]{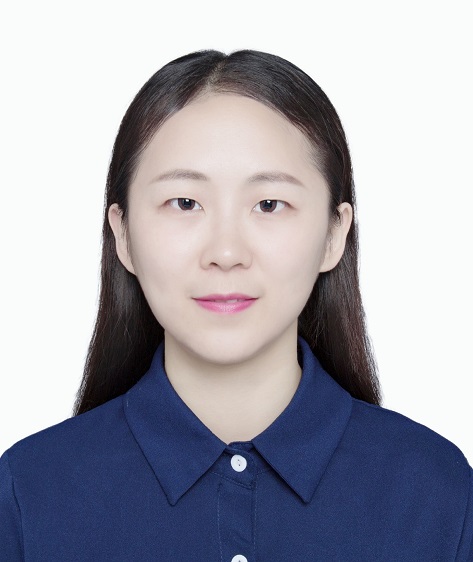}}]{Tiantian Zhang} received the B.Sc. degree in automation from the Department of Information Science and Technology, Central South University, Changsha, China, in 2015, and the M.Sc. and Ph.D. degrees in control science and engineering from the Department of Automation, Tsinghua University, Beijing, China, in 2018 and 2024, respectively. 

She is currently a Post-Doctoral Researcher with the Center for Artificial Intelligence and Robotics, Shenzhen International Graduate School, Tsinghua University. Her research interests include data science, intelligent decision-making, and reinforcement learning. She has been serving as reviewers for Nature Communications, TNNLS, TMLR and AAAI.
\end{IEEEbiography}

\begin{IEEEbiography}[{\includegraphics[width=1in,height=1.25in,clip,keepaspectratio]{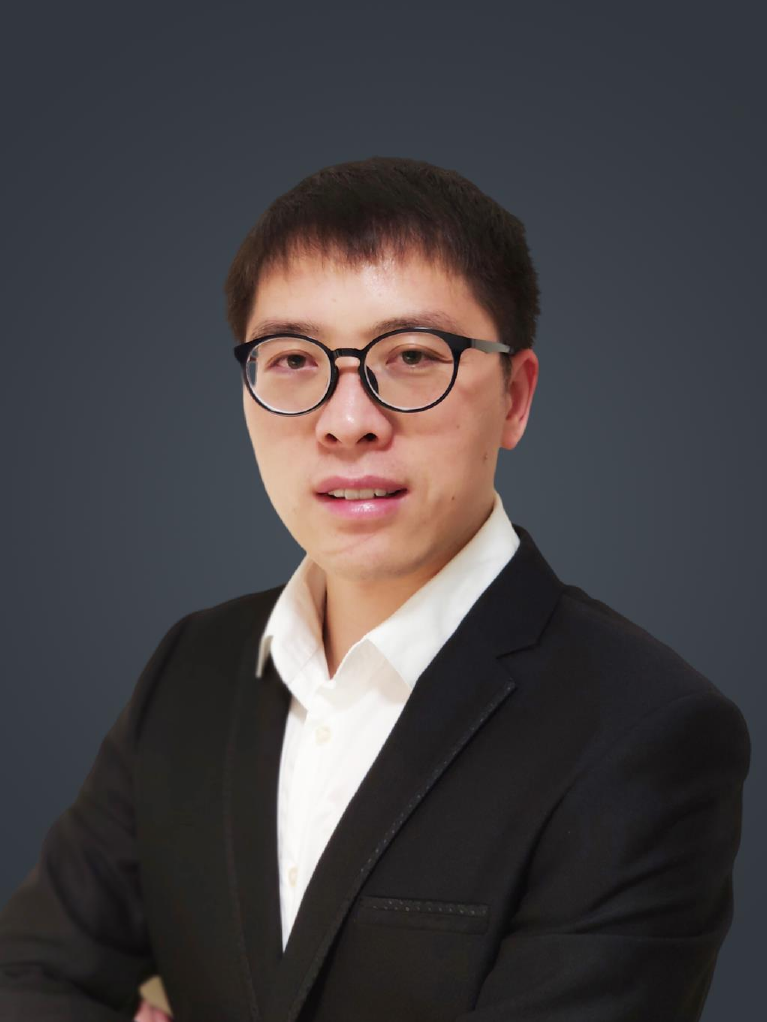}}]{Qiyue Yin} received the Ph.D. degree from the National
Laboratory of Pattern Recognition, Institute of Automation,
Chinese Academy of Sciences (CASIA), Beijing, China, in 2017. He is
currently an Associate Professor at CASIA. His major research interests include machine learning, pattern recognition and artificial intelligence on games. 
\end{IEEEbiography}

\begin{IEEEbiography}[{\includegraphics[width=1in,height=1.25in,clip,keepaspectratio]{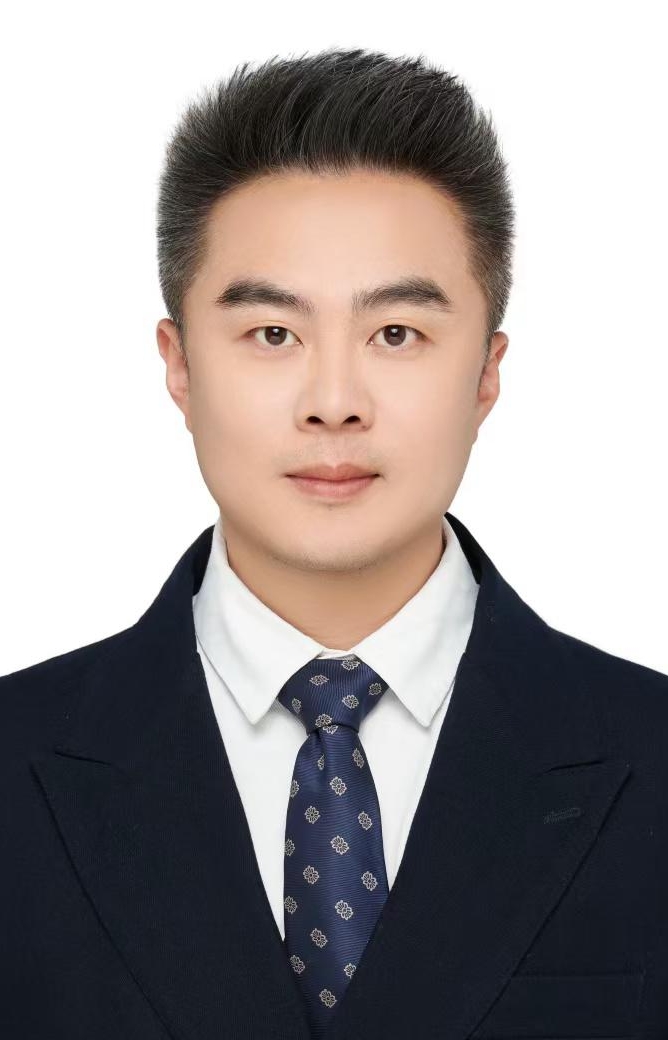}}]{Yongzhe Chang} received the Ph.D. degree in machine learning\&data analysis from University of New South Wales, Australia, in 2020. He was a post-doctoral research fellow with the AI\&Robot laboratory, Tsinghua University, from 2021 to 2023. Then he joined Tsinghua University as a associate researcher since 2023. His research interests include machine learning, reinforcement learning, large model and intelligent robot, etc.
\end{IEEEbiography}

\begin{IEEEbiography}[{\includegraphics[width=1in,height=1.25in,clip,keepaspectratio]{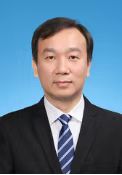}}]{Zhiheng Li} (Member, IEEE) received the Ph.D. degree in control science and engineering from Tsinghua University, Beijing, China, in 2009. He is currently an Associate Professor with the Department of Automation, Tsinghua University, and with the Graduate School at Shenzhen, Tsinghua University, Shenzhen, China. His research interests include traffic operation, advanced traffic management system, urban traffic planning, and intelligent transportation systems. Dr. Li is an Associate Editor for IEEE TRANSACTIONS ON INTELLIGENT TRANSPORTATION SYSTEMS.
\end{IEEEbiography}

\begin{IEEEbiography}[{\includegraphics[width=1in,height=1.25in,clip,keepaspectratio]{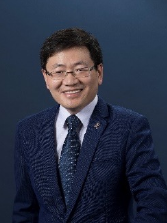}}]{Liang Wang} (Fellow, IEEE) received the B.Eng. and M.Eng. degrees from Anhui University in 1997 and 2000, respectively, and the Ph.D. degree from the Institute of Automation, Chinese Academy of Sciences (CASIA) in 2004. From 2004 to 2010, he was a research assistant with Imperial College London, United Kingdom, and Monash University, Australia, a research fellow with the University of Melbourne, Australia, and a lecturer with the University of Bath, United Kingdom, respectively. 

He is currently a full professor of the Hundred Talents Program with the National Lab of Pattern Recognition, CASIA. He has widely authored or coauthored in highly ranked international journals, such as TPAMI and TIP, and leading international conferences, such as CVPR, ICCV, and ICDM. His research interests include machine learning, pattern recognition, and computer vision. He is an IAPR fellow.
\end{IEEEbiography}

\begin{IEEEbiography}[{\includegraphics[width=1in,height=1.25in,clip,keepaspectratio]{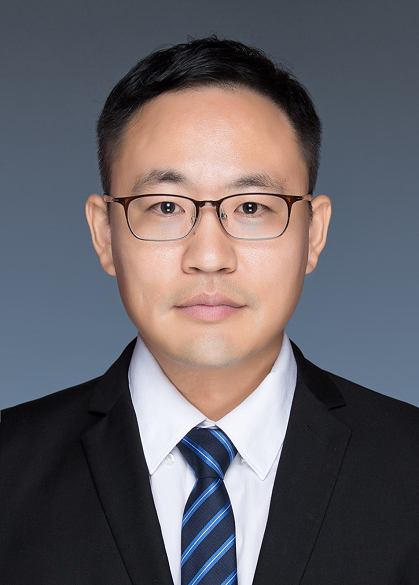}}]{Xueqian Wang} (Member, IEEE) received the M.Sc. and Ph.D. degrees in control science and engineering from the Harbin Institute of Technology (HIT), Harbin, China, in 2005 and 2010, respectively. From June 2010 to February 2014, he was a Post-Doctoral Researcher with HIT. From March 2014 to November 2019, he was an Associate Professor with the Division of Informatics, Shenzhen International Graduate School, Tsinghua University, Shenzhen, China. 

He is currently a Professor and the Leader of the Center for Artificial Intelligence and Robotics, Shenzhen International Graduate School, Tsinghua University. His research interests include robot dynamics and control, teleoperation, intelligent decision-making and game playing, and fault diagnosis.
\end{IEEEbiography}
\vspace{11pt}
\vfill

\end{document}